\documentclass[10pt,twocolumn,journal]{IEEEtran}

\usepackage{epsfig}
\usepackage{epstopdf}
\usepackage{citesort}
\usepackage{amsmath}
\usepackage{amssymb}
\usepackage{color}
\usepackage{adjustbox}
\usepackage{array}
\usepackage{multirow}
\usepackage{algorithm}
\usepackage{algorithmic}
\usepackage{rotating}
\usepackage{graphicx}
\usepackage{subfigure}
\usepackage[colorlinks,linkcolor=blue]{hyperref}
\usepackage{colortbl}
\usepackage[html]{xcolor}
\usepackage{changepage}
\usepackage{fancyhdr}
\newlength{\figurewidth}
\newlength{\smallfigurewidth}

\begin{document}

\title
{
SpectralX: Parameter-efficient Domain Generalization for Spectral Remote Sensing Foundation Models
}

\author{%
Yuxiang Zhang,
Wei Li,~\IEEEmembership{Senior Member,~IEEE},
Mengmeng Zhang, 
Jiawei Han,\\
Ran Tao,~\IEEEmembership{Senior Member,~IEEE},
Shunlin Liang,~\IEEEmembership{Fellow,~IEEE}
\thanks{%
	Y. Zhang and S. Liang are with the Jockey Club STEM Laboratory of Quantitative Remote Sensing, Department of Geography, the University of Hong Kong, Hong Kong, China (e-mail: yxzhang7@hku.hk, shunlin@hku.hk).
}
\thanks{%
	W. Li, M. Zhang, J. Han and R. Tao are with the School of Information and Electronics, Beijing Institute of Technology, and Beijing Key Laboratory of Fractional Signals and Systems, 100081 Beijing, China (e-mail: liwei089@ieee.org, mengmengzhang@bit.edu.cn, hanjiawei@bit.edu.cn, rantao@bit.edu.cn).
}
}

\maketitle
\fancyfoot[C]{\thepage}
\pagenumbering{arabic}
\renewcommand{\headrulewidth}{0pt}

\begin{abstract}
Recent advances in Remote Sensing Foundation Models (RSFMs) have led to significant breakthroughs in the field. While many RSFMs have been pretrained with massive optical imagery, more multispectral/hyperspectral data remain lack of the corresponding foundation models. To leverage the advantages of spectral imagery in earth observation, we explore whether existing RSFMs can be effectively adapted to process diverse spectral modalities without requiring extensive spectral pretraining. In response to this challenge, we proposed SpectralX, an innovative parameter-efficient fine-tuning framework that adapt existing RSFMs as backbone while introducing a two-stage training approach to handle various spectral inputs, thereby significantly improving domain generalization performance. In the first stage, we employ a masked-reconstruction task and design a specialized Hyper Tokenizer (HyperT) to extract attribute tokens from both spatial and spectral dimensions. Simultaneously, we develop an Attribute-oriented Mixture of Adapter (AoMoA) that dynamically aggregates multi-attribute expert knowledge while performing layer-wise fine-tuning. With semantic segmentation as downstream task in the second stage, we insert an Attribute-refined Adapter (Are-adapter) into the first stage framework. By iteratively querying low-level semantic features with high-level representations, the model learns to focus on task-beneficial attributes, enabling customized adjustment of RSFMs. Following this two-phase adaptation process, SpectralX is capable of interpreting spectral imagery from new regions or seasons. Comprehensive experiments across eight domain generalization tasks spanning three benchmark spectral datasets demonstrate that SpectralX effectively adapts to diverse spectral imagery and outperforms state-of-the-art methods in cross-domain interpretation tasks. The codes will be available from the website: \href{https://github.com/YuxiangZhang-BIT}{https://github.com/YuxiangZhang-BIT}.
\end{abstract}

\begin{keywords}
Multispectral image,
Hyperspectral image,
Domain generalization, 
Parameter-efficient fine-tuning,
Foundation Model
\end{keywords}

\section{Introduction}
Earth observation technology has provided profound insights into the Earth's surface by integrating diverse remote sensing data collected from various platforms and sensors. It has greatly advanced the development of key areas such as resource management, urban environmental analysis and environmental protection \cite{wu2023deep,ahn2023human,hou2024war}. With the rapid advancement of earth observation technology, vast amounts of data are continuously being released, and highly customized data present heterogeneous characteristics. Satellite platforms such as Landsat \cite{roy2014landsat}, Sentinels \cite{guanter2015enmap}, EnMAP \cite{guanter2015enmap}, and the Orbita hyperspectral micro-nano satellite (OHS) \cite{li2022whu} capture a wide range of remote sensing information, including optical RGB image, multispectral image (MSI), and hyperspectral image (HSI). These highly heterogeneous remote sensing data are difficult to be processed by a unified model.

In recent years, a series of representative self-supervised learning (SSL) techniques have been proposed in the field of computer vision, such as SimCLR \cite{chen2020simple}, Masked Autoencoders (MAE) \cite{he2022masked}, and Momentum Contrast (MoCo) \cite{he2020momentum}. These methods have significantly advanced the development of large-scale pre-trained Visual Foundation Models (VFMs). VFMs rely on large-scale unlabeled data for pre-training and are capable of learning general and universal visual representations. These representations can be efficiently transferred and adapted to downstream tasks with only a small amount of labeled data, covering applications in various fields, from image classification and semantic segmentation to object detection. Notably, the VFMs such as CLIP \cite{radford2021learning}, SAM \cite{kirillov2023segment}, and DINO \cite{caron2021emerging} have established multiple benchmarks in visual tasks. 

\begin{figure*}[tp] \small
	\vspace{-2em}
	\begin{center}
		\centering
		\epsfig{width=1.7\figurewidth,file=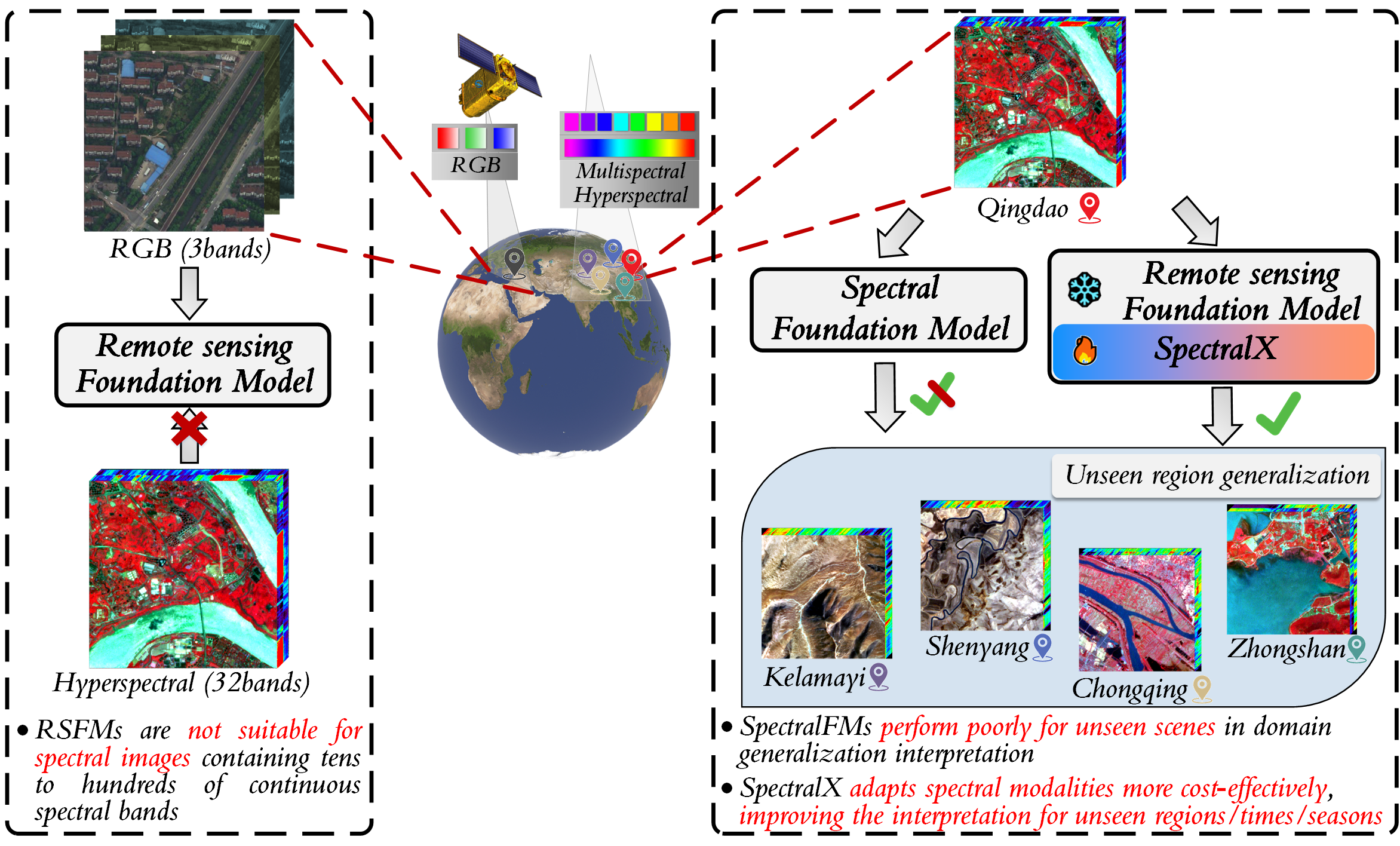}
		\caption{\label{fig:intro}
			Limitations of existing RSFMs and SpectralFMs. RGB, MSIs, and HSIs constitute diverse sources of remote sensing information. However, RSFMs pre-trained on large-scale optical modalities (with only 3 bands) are not well-suited for spectral modalities. Moreover, RSFMs and SpectralFMs often struggle when encountering unseen scenes, leading to suboptimal domain generalization performance. To address these limitations, we developed SpectralX, which effectively alleviates various spectral shifts and excels in multi-domain generalization semantic segmentation under spectral modalities.}
	\end{center}
	\vspace{-2em}
\end{figure*}

The heterogeneity of optical RGB remote sensing images and natural images is low, which makes the existing VFMs architecture suitable for the Remote Sensing Foundation Models (RSFMs). Leveraging the success of existing VFMs, several RSFMs pre-trained on optical RGB images have emerged, such as Geospatial FM (GFM) \cite{mendieta2023towards}, Scale-MAE \cite{reed2023scale}, and Cross-scale-MAE \cite{tang2024cross}. These models provide powerful general representations for remote sensing image analysis and significantly enhance the performance of downstream tasks, including classification, detection, and segmentation. Meanwhile, in order to adapt the characteristics of multispectral and hyperspectral remote sensing data, Spectral Foundation Models (SpectralFMs) designed for these data types have recently been explored. For example, SatMAE \cite{cong2022satmae}, SatMAE++ \cite{noman2024rethinking}, and SpectralGPT \cite{hong2024spectralgpt} successfully constructed SpectralFMs suitable for multispectral image analysis using less than one million multispectral Sentinel-2 remote sensing images for pre-training.
Hyperspectral data contains tens to hundreds of continuous spectral bands, and the data dimension and complexity far exceed that of RGB and multispectral data. HyperSIGMA \cite{wang2024hypersigma}, a foundation model specifically designed for hyperspectral images, collected approximately 450,000 hyperspectral images from around the world for pre-training. SpectralFMs require significantly higher training costs compared to RSFMs. On one hand, spectral data are more challenging to obtain than RGB data due to factors such as atmospheric conditions and sensor platforms. On the other hand, a single spectral image is equivalent to several RGB images, requiring more computational resources to construct SpectralFMs. To leverage the advantages of spectral imagery in earth observation through a more cost-effective manner, this presents a new challenge: \textit{\textbf{How can RSFMs be adapted to spectral modalities and achieve interpretation performance comparable to that of SpectralFMs, without requiring extensive spectral pretraining?}}

As shown in Fig.~\ref{fig:intro}, most existing RSFMs primarily focus on capturing the rich spatial information in optical RGB images, and their architectures are not suitable for spectral modalities that contain both spatial and spectral multi-dimensional information, especially for HSIs. As a result, their performance is limited when applied to downstream tasks involving spectral images. Furthermore, the currently proposed RSFMs and SpectralFMs, such as SatMAE++, SpectralGPT, and HyperSIGMA, are pre-trained on large amounts of optical and spectral data and show excellent multi-task ability in downstream task fine-tuning. However, their performance is poor when handling unseen scenes, resulting in suboptimal domain generalization performance. We constructed eight spectral domain generalization evaluation benchmarks and presented the performance of four representative models in Table \ref{table:bench}. The interpretation performance significantly degrades when dealing with unseen scenes with domain gaps. Therefore, to adapt RSFMs to spectral modalities a more economical manner and enhance their domain generalization performance in unseen scenes, we explore parameter-efficient fine-tuning (PEFT) strategies. Currently, many PEFT methods for remote sensing data design trainable adapters based on LoRA \cite{hu2022lora}, which are used to fine-tune RSFMs for specific downstream tasks \cite{peng2024parameter,10385180}. However, these methods severely overlook the inherent attributes of spectral images, limiting their adaptability on spectral image.

\begin{table*}
	\setlength\tabcolsep{0.8pt}
\caption{\label{table:bench}
	Performance benchmarking of RSFMs and SpectralFMs methods on three remote sensing datasets under eight domain generalization (DG) settings. WHUOHS \cite{li2022whu} (HSI, 32 bands) officially defines source domain (WS) and target domain (WD) with regional domain gaps. DFC2020 \cite{9369830} (MSI, 13 bands) is divided into four subsets representing seasonal domain gaps, spring, summer, autumn, and winter. MTS12 \cite{zhao2022cnn} (MSI, 6 bands, four temporal points (March, June, September, and December)) splits two regions with regional domain gaps, Slovenia northern and southern. }
	\footnotesize 
	\begin{adjustwidth}{-2cm}{-2cm}
\begin{center}
	\begin{tabular}{c|ccc|ccccc|ccc}
		\hline
		Datasets &
		\multicolumn{3}{c|}{WHUOHS (HSI)} &
		\multicolumn{5}{c|}{DFC2020 (MSI)} &
		\multicolumn{3}{c}{MTS12 (Multi-Temporal MSI)} \\ \hline
		&
		\multicolumn{1}{c|}{} &
		\multicolumn{2}{c|}{w/ regional domain gap} &
		\multicolumn{1}{c|}{} &
		\multicolumn{4}{c|}{w/ seasonal domain gap} &
		\multicolumn{1}{c|}{} &
		\multicolumn{2}{c}{w/ regional domain gap} \\ \cline{3-4} \cline{6-9} \cline{11-12} 
		\multirow{-2}{*}{DG task setting/mIoU} &
		\multicolumn{1}{c|}{\multirow{-2}{*}{\begin{tabular}{c}w/o\\ domain gap\end{tabular}}} &
		WS→WD &
		WD→WS &
		\multicolumn{1}{c|}{\multirow{-2}{*}{\begin{tabular}{c}w/o \\ domain gap\end{tabular}}} &
		Spr→S,A,W &
		Sum→S,A,W &
		Aut→S,S,W &
		Win→S,S,A &
		\multicolumn{1}{c|}{\multirow{-2}{*}{\begin{tabular}{c}w/o \\ domain gap\end{tabular}}} &
		N→S &
		S→N \\ \hline
		ScaleMAE(\textit{ICCV23}) &
		\multicolumn{1}{c|}{\cellcolor[HTML]{E7E7E7}53.42} &
		11.96(\textcolor{red}{-41.5}) &
		11.30(\textcolor{red}{-42.1}) &
		\multicolumn{1}{c|}{\cellcolor[HTML]{E7E7E7}61.87} &
		10.37(\textcolor{red}{-51.5}) &
		10.77(\textcolor{red}{-51.1}) &
		28.78(\textcolor{red}{-33.1}) &
		24.40(\textcolor{red}{-37.5}) &
		\multicolumn{1}{c|}{\cellcolor[HTML]{E7E7E7}35.94} &
		25.73(\textcolor{red}{-10.21}) &
		23.03(\textcolor{red}{-12.91}) \\
		SatMAE++(\textit{CVPR24}) &
		\multicolumn{1}{c|}{\cellcolor[HTML]{E7E7E7}55.65} &
		12.66(\textcolor{red}{-43.0}) &
		13.38(\textcolor{red}{-42.3}) &
	\multicolumn{1}{c|}{\cellcolor[HTML]{E7E7E7}63.01} &
		11.15(\textcolor{red}{-51.9}) &
		14.22(\textcolor{red}{-48.8}) &
		29.76(\textcolor{red}{-33.3}) &
		25.45(\textcolor{red}{-37.6}) &
 		\multicolumn{1}{c|}{\cellcolor[HTML]{E7E7E7}38.58} &
		25.65(\textcolor{red}{-12.93}) &
		23.82(\textcolor{red}{-14.76}) \\
		SpectralGPT+(\textit{TPAMI24}) &
		\multicolumn{1}{c|}{\cellcolor[HTML]{E7E7E7}49.33} &
		11.88(\textcolor{red}{-37.4}) &
		12.36(\textcolor{red}{-36.9}) &
		\multicolumn{1}{c|}{\cellcolor[HTML]{E7E7E7}64.90} &
		12.45(\textcolor{red}{-52.3}) &
		13.73(\textcolor{red}{-51.2}) &
		31.33(\textcolor{red}{-33.6}) &
		26.13(\textcolor{red}{-38.8}) &
		\multicolumn{1}{c|}{\cellcolor[HTML]{E7E7E7}35.95} &
		25.04(\textcolor{red}{-10.91}) &
		24.78(\textcolor{red}{-11.17})
		\\
		HyperSIGMA(\textit{arXiv24}) &
		\multicolumn{1}{c|}{\cellcolor[HTML]{E7E7E7}57.66} &
		14.08(\textcolor{red}{-43.6}) &
		14.13(\textcolor{red}{-41.5}) &
		\multicolumn{1}{c|}{\cellcolor[HTML]{E7E7E7}65.34} &
		11.17(\textcolor{red}{-54.2}) &
		13.04(\textcolor{red}{-52.3}) &
		30.64(\textcolor{red}{-34.7}) &
		28.08(\textcolor{red}{-37.3}) &	
		\multicolumn{1}{c|}{\cellcolor[HTML]{E7E7E7}41.58} &
		29.72(\textcolor{red}{-11.86}) &
		26.40(\textcolor{red}{-15.18}) \\ \hline
	\end{tabular}
	\end{center}
\end{adjustwidth}
\end{table*}


In order to address the above issues, SpectralX is proposed, a PEFT method tailored for remote sensing spectral images. Without relying on large-scale spectral pre-training, it economically adapts RSFMs to new spectral modalities and unseen scenes (unseen regions, times, and seasons). SpectralX is divided into three stages: stage1-spectral modality adaptation, stage2-task-oriented generalization training, and stage3-unseen scenes interpretation. In stage1, we focus on exploring the generalized representation of spectral images, and adopt the mask-reconstruction strategy from MAE to adapt the spectral modality. Considering the intrinsic multi-dimensional characteristics of spectral data, we explicitly decompose the spatial-spectral attributes and design a Hyper Tokenizer (HyperT) to achieve attribute perception and attribute match. Inspired by the mixture of experts (MoE), an Attribute-oriented Mixture of Adapter (AoMoA) is developed to update a small number of trainable parameters in the frozen RSFMs. The routing scheme between the attribute-specific router bank and multiple efficient tuning adapters is established. Furthermore, in stage2, we focus on task-oriented learning to capture customized features by designing an Attribute-refined Adapter (Are-adapter) that progressively refines attribute knowledge beneficial for downstream tasks. Finally, the fully trained SpectralX is used to interpret on unseen scenes in stage3. The main contributions of this paper are as follows:

\begin{itemize}

\item SpectralX, with a minimal number of trainable parameters, adapts RSFMs designed for optical modalities to spectral modalities and utilizes limited labeled data to improve the domain generalization performance of RSFMs.
\item To bridge the domain gap between optical and spectral modalities, we design the HyperT to explicitly generate tokens that capture spatial-spectral attributes.
\item AoMoA is proposed to employs flexible routing schemes for different attributes and dynamically aggregate effective expert knowledge layer by layer for updating parameters.
\item To achieve task-oriented customized adjustment, the Are-adapter is proposed. By enabling high-level tokens to continuously query low-level semantic features, it progressively refines the perception of spatial distribution and importance spectrum of land cover classes.

\end{itemize}

The rest of the paper is organized as follows. Section \ref{sec:related-works} introduces the recent research progress of Remote Sensing/Spectral Foundation Models and Parameter-Efficient Fine-Tuning. Section \ref{sec:proposed} elaborates on the proposed SpectralX. The extensive experiments and analyses on multiple spectral domain generalization evaluation benchmarks are presented in Section \ref{sec:results}. Finally, conclusions are drawn in Section \ref{sec:conclusions}.

\begin{figure*}[tp] \small
	\begin{center}
		\epsfig{width=2.2\figurewidth,file=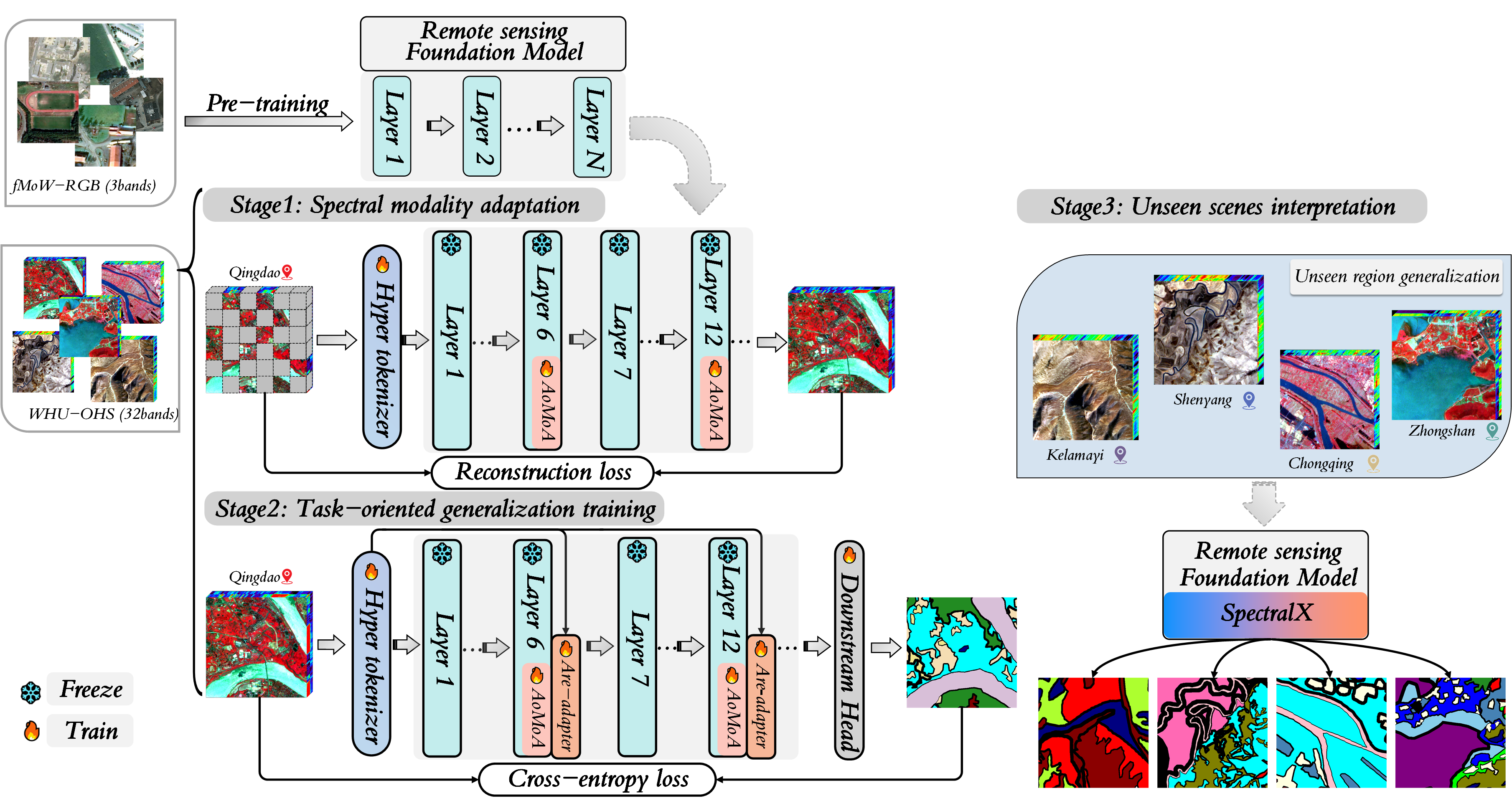}
	\end{center}
		\caption{\label{fig:SpectralX}The framework of SpectralX consists of spectral modality adaptation (stage1), task-oriented generalization training (stage2) and unseen scenes interpretation (stage3). A large-scale pre-trained RSFM for optical modalities is employed as the backbone, then it is adapted to spectral modalities through PEFT in a cost-effective manner, which could further enhance RSFM interpretation performance in domain generalization tasks, such as unseen regions and seasons.}
\end{figure*}

\begin{table}[]
	\caption{\label{table:Abbreviation}
		Summary of abbreviations.}
	\begin{center}
		\begin{tabular}{ll}
			\hline
			\multicolumn{1}{c}{Abbreviation}                      & Description                                                 \\ \hline
			MSI & Multispectral image\\
			HSI                                                   & Hyperspectral image                                     \\
			SD                                                    & Source domain \\
			TD                                                    & Target domain \\
			RSFMs                                                    & Remote Sensing Foundation Models                                             \\
			SpectralFMs                                                    & Spectral Foundation Models                                             \\
			\hline
		\end{tabular}
	\end{center}
\end{table}

\section{Related Works}
\label{sec:related-works}
\subsection{Remote Sensing/Spectral Foundation Models}

FMs leverage SSL strategies to learn generalizable representations from large-scale unlabeled data, typically without being directly tailored for specific tasks \cite{chen2024internvl,xiong2024neural,lacoste2024geo}. For downstream tasks, supervised fine-tuning with labeled data is often required to better adapt FMs to specific application requirements. MAE \cite{he2022masked} is a commonly used SSL approach where parts of the input image are randomly masked, and the model is trained to reconstruct the masked regions, thus learning the underlying visual representations. Most RSFMs and SpectralFMs are adaptations and extensions of MAE for satellite remote sensing imagery. Scale-MAE \cite{reed2023scale} applies masking to the input image at predetermined scales for pre-training. By utilizing band-pass filters, it reconstructs low- and high-frequency images at different scales, thereby exploring robust multi-scale representations of remote sensing images. Cross-Scale MAE \cite{tang2024cross} introduces scale augmentation during pre-training phase and incorporates both contrastive and generative losses to enforce cross-scale consistency learning.

Further exploring the spectral modality, SatMAE introduced temporal encoding into satellite remote sensing imagery and assigned unique spectral positional encodings to MSI, thereby enhancing the model's ability to represent spectral information. Building upon SatMAE \cite{cong2022satmae}, SatMAE++ \cite{noman2024rethinking} further supports multi-scale pre-training and incorporates a convolution-based upsampling module to reconstruct finer image details at higher scales. SpectralGPT \cite{hong2024spectralgpt} adopts a progressive training strategy designed to accommodate input data with varying sizes, resolutions, temporal sequences, and regions. By designing three-dimensional tokens, it achieves deep coupling of spatial and spectral information. Furthermore, SpectralGPT employs multi-objective reconstruction tasks to effectively capture key information within spectral sequences. HyperSIGMA \cite{wang2024hypersigma} is specifically designed for HSI and pre-trained on the large-scale hyperspectral dataset HyperGlobal-450K. It introduces a novel sparse sampling attention mechanism and a spectral enhancement module designed to integrate both spatial and spectral features, providing a more comprehensive representation of the multi-dimensional properties of HSI. Li et al. proposed a hyperspectral FM without tuning (HyperFree) \cite{li2025hyperfree}, which can process arbitrary HSIs in a prompt or zero-shot manner in different tasks. The authors collected 41,946 hyperspectral images from AVIRIS airborne sensor and designed the Hyper-Seg data engine to build nearly 150,000 images and 15.44 million masks.

\subsection{Parameter-Efficient Fine-Tuning}
Given the powerful representation capabilities of FMs, fine-tuning FMs for downstream tasks has become a popular paradigm. However, while traditional full fine-tuning is effective, it requires substantial computational and memory resources, as well as a relatively large labeled dataset. This becomes particularly expensive for FMs with billions or even trillions of parameters. Recently, PEFT has emerged as a promising solution in natural language processing (NLP) \cite{houlsby2019parameter}, addressing these challenges by limiting the number of trainable parameters during the fine-tuning stage, while potentially achieving comparable or even better performance than full fine-tuning \cite{jia2022visual,fu2024dtl}. PEFT methods train only bias parameters and introduce additional trainable adapter modules, significantly reducing the number of learnable parameters \cite{han2024parameter}. This not only facilitates more efficient adaptation to new tasks but also preserves the knowledge already learned by the FMs. Dong et al. \cite{10735106} proposed Domain-Specific and General Knowledge Fusion (DSGF) by incorporating feature-matching scores for few-shot out-of-distribution detection under different fine-tuning paradigms. Wang et al. \cite{10770155} introduced the Hyper-Adaptive network (HAda) with weight-calibrated adapters to achieve specific information transfer between multiple views. Geospatial Domain Adaptation (GDA) \cite{scheibenreif2024parameter} incorporated a Scaled Low-Rank (SLR) adapter with a small number of parameters into pre-trained RSFMs, combining it with the SSL strategy of MAE to adapt to new data modalities. Automatic Labeling for Pre-training in Segmentation (ALPS) \cite{10949707} combined Segment Anything Model and K-means to predict pseudo labels of remote sensing images without prior annotation.

\section{SpectralX}
\label{sec:proposed}
\definecolor{myblue}{RGB}{46,117,182}
\definecolor{myred}{RGB}{192,0,0}
\definecolor{mypurple}{RGB}{135,98,232}

\begin{figure*}[tp] \small
	\vspace{-2em}
	\begin{center}
		\centering
		\epsfig{width=2\figurewidth,file=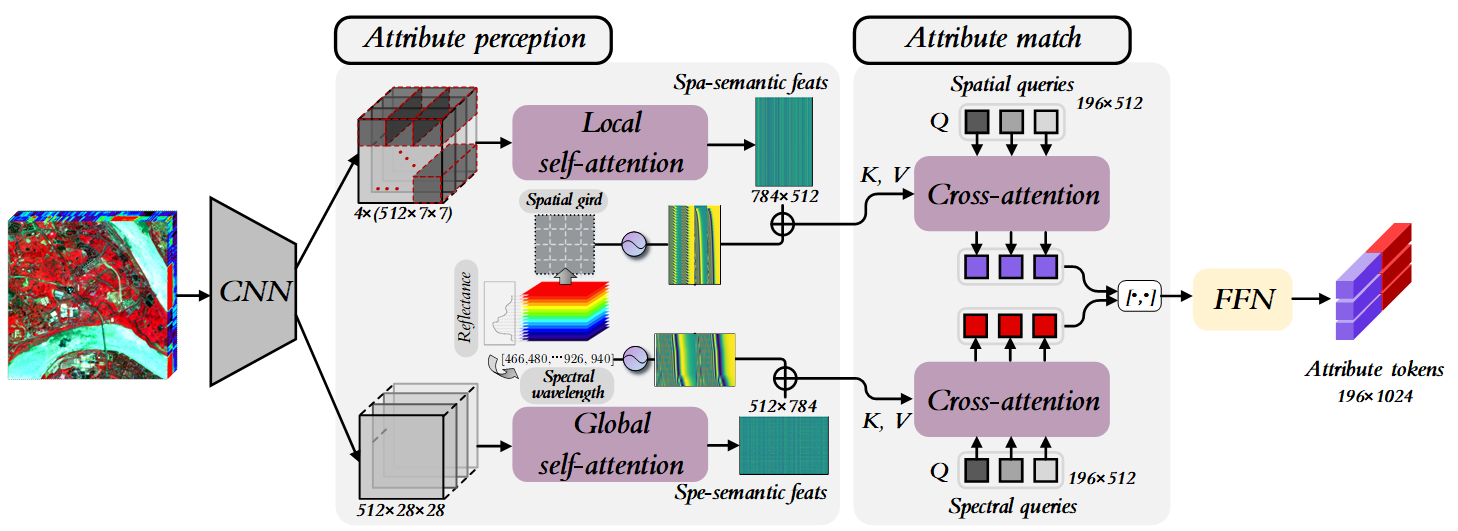}
		\caption{\label{fig:HyperT}The flowchart of HyperT. A lightweight CNN is employed to downsample the spectral image. The downsampled features are then fed into the attribute perception module, which utilizes local and global self-attention to explore spatial and spectral attributes, respectively. Specifically, the spatial grid coordinates and spectral wavelengths are explicitly incorporated to compute position embeddings, which are added to the semantic features before being input to the attribute matching module. In the attribute matching module, predefined queries are split along the channel dimension—the first half for spatial attribute matching and the second half for spectral attribute matching. The representations are then merged to generate the final attribute tokens.}
	\end{center}
	\vspace{-2em}
\end{figure*}
In this section, we first introduce the overall framework of SpectralX in \ref {subsec:Overview}. The details of HyperT, AoMoA, and Are-Adapter are presented in \ref {subsec:tokenizer}, \ref {subsec:AoMoA}, \ref {subsec:Are}, respectively. 
\subsection{Overview of SpectralX}
\label{subsec:Overview}

Abbreviations used in this paper is summarized in Table \ref{table:Abbreviation}. Assume that ${{\bf{X}}}\in\mathbb{R}{^{H \times W \times d}}$ is the data from source domain, and ${{\bf{Y}}}\in\mathbb{R}{^{H \times W}}$ is the corresponding segmentation maps (RSFMs default input image space size $H \times W$ is 224$\times$224). Here, $d$ and $N$ denote the dimension of data and the number of source samples, respectively.

The main framework of proposed SpectralX is composed of three stages (Fig.~\ref{fig:SpectralX}): spectral modality adaptation (stage1), task-oriented generalization training (stage2) and unseen scenes interpretation (stage3). In stage1, the training objective is the mask-reconstruction task. We use RSFMs (Scale-MAE \cite{reed2023scale} and SatMAE++ \cite{noman2024rethinking}) pre-trained on large-scale unlabeled optical modality data as the backbone, freezing all layer parameters and only updating a small number of parameters in the HyperT (Fig.~\ref{fig:HyperT}) and AoMoA (Fig.~\ref{fig:AoMoA}). First, the spectral modality data is input into the HyperT to achieve a preliminary transformation from the optical modality to the spectral modality at the representation level. Through the attention mechanism, the attributes of both spatial and spectral dimensions are perceived separately, generating tokens with spatial-spectral attributes suitable for the spectral modality. Furthermore, AoMoA is inserted into 4 specific transformer blocks in both the encoder and decoder for efficient fine-tuning of RSFM and modality adaptation. AoMoA consists of an attribute-specific router bank and an attribute-shared adapter bank. In stage2, the training objective is the downstream task (segmentation). Building upon the model architecture of stage1 (discarding the decoder), we plug in the Are-adapter (Fig.~\ref{fig:Are}) after the transformer blocks of encoder with AoMoA, and use the UperNet \cite{xiao2018unified} as the downstream head. Are-adapter takes as input the high-level attribute token representations from the previous layer and the low-level semantic features obtained by the HyperT. Through continuously querying the potential spatial-spectral attribute knowledge of low-level features through high-level tokens, a match map is obtained. This match map is then utilized to extract importance attribute tokens and improve attribute perception capability. After completing the training of these two stages, SpectralX is used to interpret unseen scenes in stage3. As shown in Fig.~\ref{fig:SpectralX}, Qingdao (seen region) is used for training in stage1 and stage2, while Kelamayi, Shenyang, Chongqing, and Zhongshan (unseen regions) are interpreted in stage3.

In the training phase, the loss functions are defined as follows:  
\begin{itemize}
\item In stage1, the training procedure follows MAE, utilizing the reconstruction loss ${{\cal L}_{mae}}$. 
\item In stage2, for the segmentation task, the cross-entropy loss \({{\cal L}_{ce}}\) is adopted.
\end{itemize}

\subsection{Hyper tokenizer (HyperT)}
\label{subsec:tokenizer}

Spectral modality and optical modality inherently exhibit heterogeneity. The most straightforward approach to adapt transformer-based RSFMs to spectral modality is to replace the original projection layer (which projects data into token embeddings suitable for transformer input) and fine-tune the model. However, this is not the optimal method for modality adaptation. To more effectively mitigate the domain gap between the two modalities, we designed the HyperT to replace the projection layer in RSFMs, with details illustrated in Fig.~\ref{fig:HyperT}. 

First, the training images are input into a CNN composed of multiple Conv2d-BN2d-GELU (BN2d-BatchNorm2d) layers for downsampling. And then, in the attribute perception module, we introduced the local self-attention and global self-attention from SAM \cite{kirillov2023segment} to decouple the spatial and spectral dimensions of the spectral images. The local self-attention perceives spatial attributes (spatial local details) to obtain spa-semantic features ${{\bf{Z}}_{spa}} \in\mathbb{R}{^{784 \times 512}}$, while the global self-attention perceives spectral attributes (long-range dependencies between spectral bands) to obtain spe-semantic features ${{\bf{Z}}_{spe}} \in\mathbb{R}{^{512 \times 784}}$. Furthermore, the spatial grid and the spectral wavelength (e.g. the wavelength of DFC2020 dataset ${[443nm,490nm,560nm,...,1610nm,2190nm]}$) are utilized to compute position embeddings that match the shapes of the two types of semantic features. Here, since the number of bands in the original spectral image is much smaller than the number of spe-semantic features, we generate one-dimensional sine-cosine position embeddings based on the original spectral wavelengths and then perform interpolation. These two types of position embeddings are added to the semantic features and serve as the K (keys) and V (values) inputs to the attribute match module. 

We need to obtain a compact vocabulary set of size $L$ (i.e., attribute tokens ${{\bf{T}}_{att}} \in\mathbb{R}{^{196 \times 1024}}$, $L$=196) through the attribute match module. In order to make tokens capture spatial-spectral attributes, we partition along the channel dimension, assigning the first 512 dimensions to spatial attributes and the latter 512 dimensions to spectral attributes. Specifically, we initialize the Q (queries) for both spatial and spectral attributes and perform cross-attention operations with their corresponding K and V. The results are then concatenated and fed into a Feed-Forward Network (FFN) to obtain the ${{\bf{T}}_{att}}$.

\subsection{Attribute-oriented Mixture of Adapter (AoMoA)}
\label{subsec:AoMoA}

\begin{figure}[tp] \small
	\begin{center}
		\centering
		\epsfig{width=1\figurewidth,file=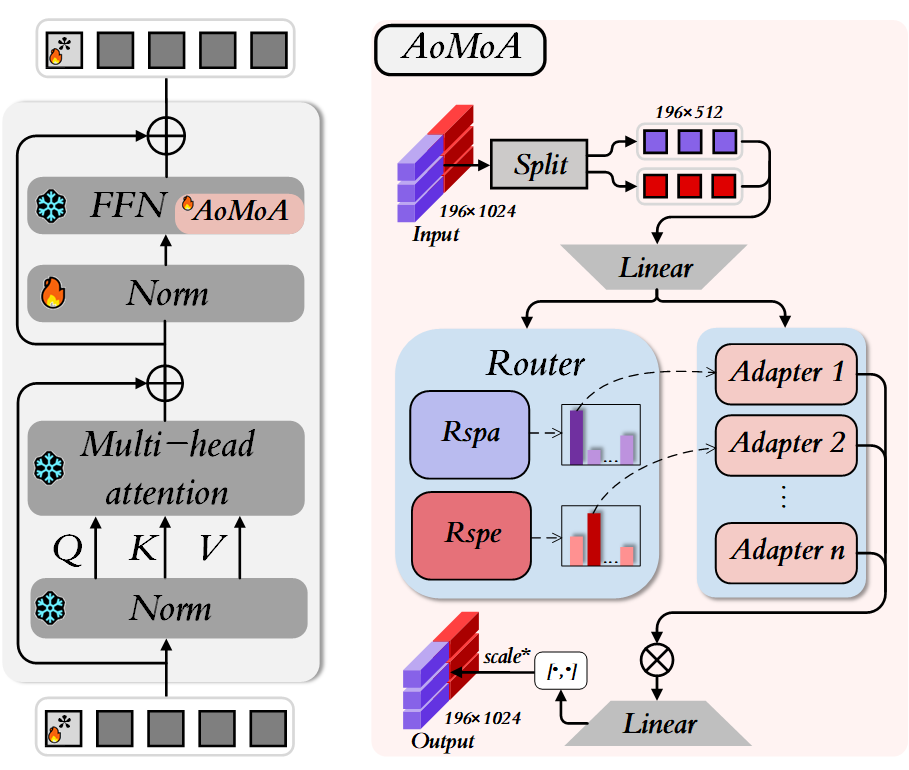}
		\caption{\label{fig:AoMoA}
			The flowchart of AoMoA. The attribute token representation is input and split along the channel dimension. The attribute-specific router is used to adaptively select adapters to process its corresponding attribute tokens. The output is a token representation of the same shape as the input. The instances of AoMoA is inserted into the FFN layer of 4 specific transformer blocks in both the encoder and decoder.}
	\end{center}
\end{figure}

HyperT takes into full consideration the characteristics of spectral images at the input stage. However, merely crafting well-fitted tokens is insufficient to adapt RSFMs to the spectral modality. We propose AoMoA for efficiently fine-tuning RSFMs. During this process, AoMoA dynamically allocates useful attribute knowledge from the token representations, enabling effective modality adaptation. Drawing inspiration from the general structure of MoA, AoMoA consists of a down-projection matrix \( {\bf{W}}_{down} \in\mathbb{R}{^{r \times (r/4)}} \), followed by an attribute-specific router bank $\left\{ {{{\cal{R}}_{spa}},{{\cal{R}}_{spe}}} \right\}$, an attribute-shared adapter bank $\left\{ {{{\cal{A}}_{1}},{{\cal{A}}_{2}},...,{{\cal{A}}_{N_a}}} \right\}$, and an up-projection matrix \( {\bf{W}}_{up} \in\mathbb{R}{^{(r/4) \times r}} \)  ($N_a$=4, $r$=512). As illustrated in Fig.~\ref{fig:AoMoA}, the input attribute token representations ${{\bf{T}}_{att}}$ are split along the channel dimension (${{\bf{T}}_{spa}} \in\mathbb{R}{^{196 \times 512}}$ and ${{\bf{T}}_{spe}} \in\mathbb{R}{^{196 \times 512}}$) and multiplied by ${\bf{W}}_{down}$ (a linear layer) to reduce dimensionality. The following uses ${{\bf{T}}_{spa}}$ as an example to describe the details of AoMoA.

We select the spatial attribute-specific router ${{\cal{R}}_{spa}}$ from the router bank to customize the routing scheme, determining which adapters in the adapter bank should process the spatial attribute tokens. The routing values is calculated as follows,
\begin{equation}\label{eq:1}
\begin{split}
\mathcal{R}_{spa}(\mathbf{T}_{spa}) = softmax\bigl( & TopK(\mathbf{T}_{spa}\mathbf{W}^{0}_{gate}) \\
& + \mathcal{N}(0,1) \cdot softplus(\mathbf{T}_{spa}\mathbf{W}^{0}_{noise}) \bigr)
\end{split}
\end{equation}
where \({{\bf{W}}_{gate}} \in\mathbb{R}{^{2\times(r/4)\times{N_a}}}\) represents the learnable gating values, which determine which adapters the current token should be assigned to. \({{\bf{W}}_{noise}} \in\mathbb{R}{^{2\times(r/4)\times{N_a}}}\) represents the learnable noise. Their first dimension corresponds to spatial attribute and spectral attribute. During training phase, noise is added to \({{\bf{W}}_{gate}}\) to introduce a certain level of randomness, thereby preventing certain adapters from being over-selected or ignored. \(TopK \left( \cdot \right)\) retains only the top K (where K = 2) values. Based on the obtained routing values \({{\cal{R}}_{spa}}\left( {{{\bf{T}}_{spa}}} \right)\), an appropriate mixture of adapters is customized as follows,
\begin{equation}\label{eq:2}
{{{\bf{\hat T}}}_{spa}} = \sum\limits_{i = 0}^{{N_a}} {{{\cal{R}}_{spa}}{{\left( {{{\bf{T}}_{spa}}} \right)}_i}{{\cal{A}}_i{{\left( {{{\bf{T}}_{spa}}} \right)}}}}
\end{equation}
where adapters are constructed using multi-layer perceptrons (MLPs). Different attributes are routed to different adapters to achieve dynamic aggregation of expert knowledge. The aggregation features of spectral attributes are obtained by using similar routing policies as described above. The aggregation features of spectral attributes C are obtained by using similar routing scheme as described above. Subsequently, \( {{{\bf{\hat T}}}_{spa}} \) and \( {{{\bf{\hat T}}}_{spe}} \) are concatenated, multiplied by \( {\bf{W}}_{up} \) (a linear layer), and then multiplied by a learnable scaling vector ${\bf{s}}_{1} \in\mathbb{R}{^{2r}}$ to obtain the output of AoMoA,
\begin{equation}\label{eq:3}
{f_{AoMoA}} = {{\bf{s}}_1} \left( {\left[ {{\bf{W}}_{up}}{{{{\bf{\hat T}}}_{spa}};{{\bf{W}}_{up}}{{{\bf{\hat T}}}_{spe}}} \right]} \right){\rm{ }}
\end{equation}
In the experiment, we used the ViT-large model (which includes 24 transformer blocks) as the backbone. AoMoA is inserted into 4 specific transformer blocks in both the encoder and decoder (the 6th, 12th, 18th, and 24th layers). Specifically, as shown in Fig.~\ref{fig:AoMoA}, we added instances of AoMoA to the FFN layer of the transformer block, using \( f_{AoMoA} \) to fine-tune the features output by the FFN layer,
\begin{equation}\label{eq:4}
{{\hat f}_{FFN}} = {f_{FFN}} + {f_{AoMoA}}
\end{equation}

\begin{figure}[tp] \small
	\begin{adjustwidth}{-0.5cm}{0cm}
		\begin{center}
			\centering
			\epsfig{width=1.2\figurewidth,file=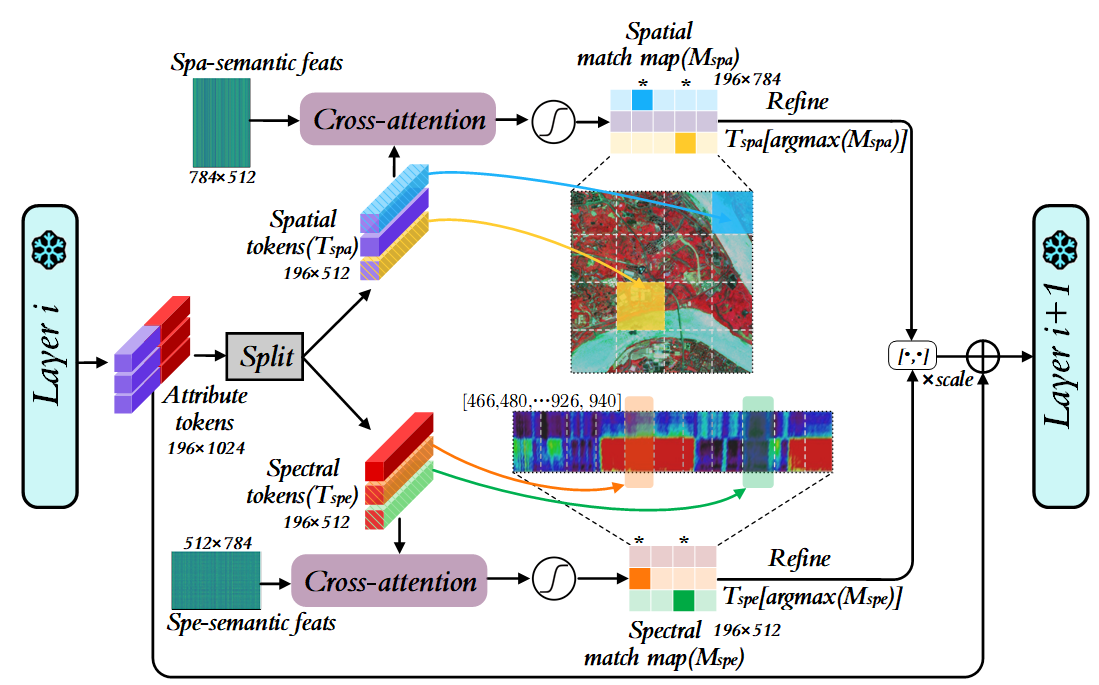}
		\end{center}
	\end{adjustwidth}
			\caption{\label{fig:Are}
	The flowchart of Are-adapter. The attribute token representations and the semantic features output by the HyperT are taken as inputs. Two attribute match map are generated by querying low-level semantic features, which is utilized to select and refine the most relevant attribute tokens for each semantic feature. The Are-adapter is applied only in stage2, where it is inserted into the transformer blocks of the encoder alongside AoMoA.}
\end{figure}

\subsection{Attribute-refined Adapter (Are-adapter)}
\label{subsec:Are}

In the training of spectral modality adaptation (stage1), the transition from the optical modality to the spectral modality is accomplished, while general spectral features are learned. Next, we focus on the generalization training of SpectralX for downstream tasks (segmentation) in stage2. There exists a task discrepancy between stage1 and stage2, namely the gap between masked image reconstruction and semantic segmentation tasks. The Are-adapter is proposed to mitigate this task gap and achieve task-oriented customization, as illustrated in Fig.~\ref{fig:Are}. The Are-adapter takes two types of inputs: attribute token representations \({{\bf{T}}_{att}}\) (decomposed into \({{\bf{T}}_{spa}}\) and \({{\bf{T}}_{spe}}\)), and semantic features \({{\bf{Z}}_{spa}}\) and \({{\bf{Z}}_{spe}}\) obtained by the HyperT. To guide the model to consistently focus on the spatial distribution and the importance of spectral bands of land cover classes, we employ cross-attention to generate match maps \({{\bf{M}}_{spa}} \in\mathbb{R}{^{196 \times 784}}\) and \({{\bf{M}}_{spe}} \in\mathbb{R}{^{196 \times 512}}\), which capture the associations between semantic features and attribute tokens. Based on the match maps, the most relevant attribute token for each semantic feature is selected and refined,
\begin{equation}\label{eq:5}
\begin{array}{l}
{{{\bf{T'}}}_{spa}} = refin{e_1}\left( {{{\bf{T}}_{spa}}\left[ {argmax\left( {{{\bf{M}}_{spa}}} \right)} \right]} \right)\\
{{{\bf{T'}}}_{spe}} = refin{e_2}\left( {{{\bf{T}}_{spe}}\left[ {argmax\left( {{{\bf{M}}_{spe}}} \right)} \right]} \right)
\end{array}
\end{equation}
where, \(refin{e_1}\) is designed for the spatial dimension and consists of LayerNorm-Conv2d-BN2d-ReLU (need to reshape features). \(refin{e_2}\) is designed for the spectral dimension and consists of LayerNorm-Conv1d-BN1d-ReLU. The refined attribute tokens \({{{\bf{T'}}}_{spa}}\) and \({{{\bf{T'}}}_{spe}}\) are concatenated and multiplied by a learnable scaling vector ${\bf{s}}_{2} \in\mathbb{R}{^{2r}}$ to adjust the attribute token \({{\bf{T}}_{att}}\),
\begin{equation}\label{eq:6}
{{\bf{T'}}_{att}} = {{\bf{T}}_{att}} + {{\bf{s}}_2} \cdot \left( {\left[ {{{{\bf{T'}}}_{spa}};{{{\bf{T'}}}_{spe}}} \right]} \right)
\end{equation}
Note that the decoder is removed in stage2, and Are-adapter is only inserted into the transformer blocks of the encoder with AoMoA.

\section{Experimental results and analysis}
\label{sec:results}

\begin{table*}[]
	\setlength\tabcolsep{2.5pt}
\caption{\label{table:data}
	The task abbreviations, the number of training and test images, image sizes, and the number of classes of WHUOHS \cite{li2022whu}, DFC2020 \cite{9369830} and MTS12 \cite{zhao2022cnn}. }
	\footnotesize
	\begin{center}
	\begin{tabular}{cccccccccc}
		\hline 
		Dataset &
		Band &
		Domain gap &
		Source domain &
		Target domain &
		Abbreviation &
		Train number &
		Test number &
		Image size &
		Classes \\ \hline
		\multirow{3}{*}{WHUOHS} &
		32 &
		w/o domain gap &
		- &
		- &
		- &
		4822 &
		2460 &
		512$\times$512 &
		24 \\ \cline{2-10} 
		&
		\multirow{2}{*}{32} &
		\multirow{2}{*}{Unseen region} &
		WHU-SD &
		WHU-TD &
		WS→WD &
		1464 &
		1450 &
		512$\times$512 &
		22 \\
		&
		&
		&
		WHU-TD &
		WHU-SD &
		WD→WS &
		1450 &
		1464 &
		512$\times$512 &
		22 \\ \hline
		\multirow{5}{*}{DFC2020} &
		13 &
		w/o domain gap &
		- &
		- &
		- &
		4279 &
		1835 &
		256$\times$256 &
		8 \\ \cline{2-10} 
		&
		\multirow{4}{*}{13} &
		\multirow{4}{*}{Unseen season} &
		Spring &
		Summer\&Autumn\&Winter &
		Spr→S,A,W &
		1444 &
		4670 &
		256$\times$256 &
		8 \\
		&
		&
		&
		Summer &
		Spring\&Autumn\&Winter &
		Sum→S,A,W &
		486 &
		5628 &
		256$\times$256 &
		8 \\
		&
		&
		&
		Autumn &
		Spring\&Summer\&Winter &
		Aut→S,S,W &
		3016 &
		3098 &
		256$\times$256 &
		8 \\
		&
		&
		&
		Winter &
		Spring\&Summer\&Autumn &
		Win→S,S,A &
		1168 &
		4946 &
		256$\times$256 &
		8 \\ \hline
		\multirow{3}{*}{MTS12} &
		24 &
		w/o domain gap &
		- &
		- &
		- &
		655 &
		281 &
		500$\times$500 &
		8 \\ \cline{2-10} 
		&
		\multirow{2}{*}{24} &
		\multirow{2}{*}{Unseen region} &
		Northern &
		Southern &
		N→S &
		639 &
		297 &
		500$\times$500 &
		8 \\
		&
		&
		&
		Southern &
		Northern &
		S→N &
		297 &
		639 &
		500$\times$500 &
		8 \\ \hline 
	\end{tabular}
	\end{center}
\end{table*}

\begin{table*}[]
	\setlength\tabcolsep{1.5pt}
\caption{\label{table:Ablation}
	Ablation studies of key components of SpectralX on the WHUOHS and DFC2020 datasets, shown by mIoU. }
\begin{center}
	\begin{tabular}{c|cc|c|ccc|ccccc}
		\hline
		\multirow{3}{*}{Backbone} &
		\multicolumn{2}{c|}{\multirow{3}{*}{Ablation setting}} &
		\multirow{3}{*}{\begin{tabular}[c]{c}Trainable\\ params\end{tabular}} &
		\multicolumn{3}{c|}{WHUOHS} &
		\multicolumn{5}{c}{DFC2020} \\ \cline{5-12} 
		&
		\multicolumn{2}{c|}{} &
		&
		\multicolumn{1}{c|}{\multirow{2}{*}{\begin{tabular}[c]{c}w/o\\ domain gap\end{tabular}}} &
		\multicolumn{2}{c|}{w/ regional domain gap} &
		\multicolumn{1}{c|}{\multirow{2}{*}{\begin{tabular}[c]{c}w/o\\ domain gap\end{tabular}}} &
		\multicolumn{4}{c}{w/ seasonal domain gap} \\ \cline{6-7} \cline{9-12} 
		&
		\multicolumn{2}{c|}{} &
		&
		\multicolumn{1}{c|}{} &
		WS→WD &
		WD→WS &
		\multicolumn{1}{c|}{} &
		Spr→S,A,W &
		Sum→S,A,W &
		Aut→S,S,W &
		Win→S,S,A \\ \hline
		\multirow{8}{*}{ScaleMAE} &
		\multicolumn{2}{c|}{Full fine-tuning} &
		303.4M &
		\multicolumn{1}{c|}{53.4} &
		12.0 &
		11.3 &
		\multicolumn{1}{c|}{61.9} &
		10.4 &
		10.8 &
		28.8 &
		24.4 \\ \cline{2-12} 
		&
		\multicolumn{2}{c|}{Freeze} &
		3.4M &
		\multicolumn{1}{c|}{48.2} &
		9.8 &
		9.3 &
		\multicolumn{1}{c|}{57.5} &
		7.6 &
		7.3 &
		25.1 &
		20.8 \\ \cline{2-12} 
		&
		\multicolumn{1}{c|}{\multirow{2}{*}{\begin{tabular}[c]{c}w/o\\ stage1\end{tabular}}} &
		w/ HyperT &
		0.4M &
		\multicolumn{1}{c|}{49.5} &
		10.2 &
		9.7 &
		\multicolumn{1}{c|}{59.6} &
		9.1 &
		9.6 &
		27.8 &
		23.6 \\
		&
		\multicolumn{1}{c|}{} &
		+ w/ AoMoA &
		8.2M &
		\multicolumn{1}{c|}{55.1} &
		12.9 &
		12.0 &
		\multicolumn{1}{c|}{63.5} &
		12.0 &
		12.7 &
		30.2 &
		26.8 \\
		&
		\multicolumn{1}{c|}{} &
		+ w/ Are-adapter &
		11.8M &
		\multicolumn{1}{c|}{56.4} &
		13.8 &
		13.1 &
		\multicolumn{1}{c|}{64.4} &
		12.8 &
		13.3 &
		31.5 &
		27.5 \\ \cline{2-12} 
		&
		\multicolumn{1}{c|}{\multirow{2}{*}{\begin{tabular}[c]{c}w/\\ stage1\end{tabular}}} &
		w/ HyperT &
		0.4M &
		\multicolumn{1}{c|}{51.7} &
		11.4 &
		10.9 &
		\multicolumn{1}{c|}{61.5} &
		10.2 &
		10.4 &
		28.5 &
		24.1 \\
		&
		\multicolumn{1}{c|}{} &
		+ w/ AoMoA &
		8.2M &
		\multicolumn{1}{c|}{58.8} &
		14.5 &
		13.8 &
		\multicolumn{1}{c|}{65.9} &
		13.5 &
		13.9 &
		31.7 &
		27.8 \\
		&
		\multicolumn{1}{c|}{} &
		+ w/ Are-adapter &
		11.8M &
		\multicolumn{1}{c|}{\textbf{59.7}} &
		\textbf{15.6} &
		\textbf{14.9} &
		\multicolumn{1}{c|}{\textbf{66.7}} &
		\textbf{13.9} &
		\textbf{14.6} &
		\textbf{32.3} &
		\textbf{28.6} \\ \hline
	\end{tabular}
	\end{center}
\end{table*}

\begin{table*}[]
		\setlength\tabcolsep{1pt}
	\caption{\label{table:WHUOHS_miou}
		Performance comparison without domain gap on WHUOHS. Bold are the best scores, underline are second ones. }
	\begin{adjustwidth}{-0.5cm}{-0.5cm}
	\begin{tabular}{cccccccccccccccccccccccccccc}
		\hline
		Method &
		Backbone &
		\begin{tabular}[c]{c}Trainable\\ params\end{tabular} &
		C1 &
		C2 &
		C3 &
		C4 &
		C5 &
		C6 &
		C7 &
		C8 &
		C9 &
		C10 &
		C11 &
		C12 &
		C13 &
		C14 &
		C15 &
		C16 &
		C17 &
		C18 &
		C19 &
		C20 &
		C21 &
		C22 &
		C23 &
		C24 &
		mIoU \\ \hline
		\multicolumn{28}{c}{\cellcolor[HTML]{E7E7E7}\textit{\textbf{Domain Adaptation/Generalization}}} \\ \hline
		SSC &
		ResNet101 &
		50.4M &
		70.4 &
		69.1 &
		62.5 &
		38.6 &
		0.63 &
		0.0 &
		47.8 &
		32.1 &
		37.2 &
		61.3 &
		89.5 &
		47.9 &
		0.0 &
		47.3 &
		75.1 &
		\underline {47.6} &
		24.5 &
		69.4 &
		71.1 &
		9.6 &
		26.1 &
		0.0 &
		72.7 &
		95.9 &
		45.7 \\
		DAEN &
		RefineNet-lw &
		48.7M &
		70.1 &
		70.8 &
		64.8 &
		40.3 &
		13.3 &
		4.3 &
		48.7 &
		32.3 &
		40.5 &
		62.0 &
		85.7 &
		52.5 &
		48.7 &
		54.8 &
		76.6 &
		53.6 &
		34.5 &
		76.5 &
		69.9 &
		31.4 &
		48.3 &
		42.9 &
		76.9 &
		97.2 &
		54.0 \\
		DSTC &
		PVTv2-B1 &
		25.7M &
		\underline {71.5} &
		71.3 &
		63.7 &
		37.9 &
		\underline {17.9}&
		19.7 &
		\underline{56.3} &
		\underline {36.2} &
		\underline {45.8} &
		\underline {70.5} &
		93.3 &
		\underline {61.4} &
		73.3 &
		57.9 &
		\underline {78.0} &
		47.2 &
		\underline {40.1} &
		78.9 &
		\underline {75.3} &
		\underline {63.0} &
		\underline {53.8} &
		39.8 &
		\underline {78.7} &
		97.1 &
		54.5 \\ \hline
		\multicolumn{28}{c}{\cellcolor[HTML]{E7E7E7}\textit{\textbf{Remote Sensing/Spectral Foundation Models}}} \\ \hline
		ScaleMAE$^{\textbf{\dag}}$ &
		ViT-L &
		303.4M &
		66.8 &
		65.8 &
		61.6 &
		38.8 &
		16.9 &
		18.4 &
		52.8 &
		32.8 &
		38.7 &
		67.3 &
		85.4 &
		54.0 &
		49.9 &
		56.7 &
		71.3 &
		42.3 &
		32.0 &
		77.0 &
		67.2 &
		36.3 &
		41.3 &
		\underline {45.8} &
		72.2 &
		90.7 &
		53.4 \\
		SatMAE++$^{\textbf{\dag}}$ &
		ViT-L &
		305.5M &
		\textbf{72.0} &
		\textbf{72.5} &
		\textbf{66.6} &
		\textbf{42.4} &
		19.5 &
		\underline {20.4} &
		54.6 &
		34.9 &
		39.5 &
		69.4 &
		87.6 &
		56.1 &
		38.2 &
		58.0 &
		75.8 &
		51 &
		38.5 &
		79.0 &
		65.6 &
		40.0 &
		45.9 &
		39.3 &
		73.0 &
		95.9 &
		55.7 \\
		SpectralGPT+$^{\textbf{\dag}}$ &
		ViT-L &
		307.2M &
		63.4 &
		63.0 &
		58.7 &
		32.6 &
		17.2 &
		14.4 &
		50.1 &
		29.1 &
		39.1 &
		52.8 &
		84.5 &
		44.8 &
		39.2 &
		49.0 &
		69.4 &
		37.7 &
		31.5 &
		70.4 &
		65.7 &
		54.7 &
		35.1 &
		36.2 &
		56.8 &
		88.6 &
		49.3 \\
		HyperSIGMA$^{\textbf{\dag}}$ &
		ViT-L &
		382.5M &
		71.2 &
		\underline {71.4} &
		\underline {65.0} &
		\underline {40.4} &
		17.6 &
		19.1 &
		\textbf{56.9} &
		35.5 &
		\textbf{46.6} &
		66.6 &
		92.0 &
		58.5 &
		59.5 &
		58.0 &
		75.0 &
		\textbf{48.5} &
		37.4 &
		\underline {79.1} &
		74.1 &
		39.6 &
		51.1 &
		45.5 &
		77.5 &
		97.5 &
		57.7 \\ \hline
		\multicolumn{28}{c}{\cellcolor[HTML]{E7E7E7}\textit{\textbf{Parameter-Efficient Fine-Tuning}}} \\ \hline
		LoRA\tiny $*$ScaleMAE &
		ViT-L &
		6.9M &
		68.5 &
		69.0 &
		62.8 &
		35.4 &
		17.4 &
		17.6 &
		54.0 &
		31.4 &
		40.6 &
		66.4 &
		75.4 &
		52.9 &
		51.9 &
		54.9 &
		76.7 &
		45.7 &
		37.7 &
		76.6 &
		57.7 &
		24.6 &
		44.2 &
		39.7 &
		63.8 &
		85.6 &
		52.1 \\
		LoRA\tiny $*$SatMAE++ &
		ViT-L &
		6.9M &
		67.0 &
		66.2 &
		60.3 &
		35.9 &
		16.7 &
		11.4 &
		49.8 &
		29.0 &
		38.9 &
		62.2 &
		74.6 &
		55.2 &
		24.9 &
		\underline {58.3} &
		73.8 &
		41.3 &
		30.7 &
		76.4 &
		52.8 &
		28.2 &
		41.6 &
		23.7 &
		67.1 &
		79.0 &
		48.6 \\
		SLR\tiny $*$ScaleMAE &
		ViT-L &
		7.3M &
		69.4 &
		69.2 &
		62.1 &
		32.4 &
		17.7 &
		16.4 &
		54.8 &
		30.6 &
		40.9 &
		66.6 &
		77.1 &
		56.9 &
		45.3 &
		58.0 &
		75.9 &
		44.8 &
		37.8 &
		73.6 &
		59.3 &
		34.2 &
		50.3 &
		35.6 &
		62.9 &
		85.3 &
		52.4 \\
		SLR\tiny $*$SatMAE++ &
		ViT-L &
		7.3M &
		66.6 &
		66.2 &
		59.8 &
		38.1 &
		13.5 &
		13.4 &
		54.1 &
		29.8 &
		39.2 &
		63.0 &
		70.5 &
		48.1 &
		32.9 &
		55.6 &
		73.6 &
		41.4 &
		35.4 &
		74.3 &
		51.3 &
		13.5 &
		48.3 &
		23.3 &
		62.8 &
		77.1 &
		48.0 \\
		SpectralX\tiny $*$ScaleMAE &
		ViT-L &
		11.8M &
		\underline{71.5} &
		71.3 &
		63.7 &
		37.9 &
		\underline {17.9} &
		19.7 &
		\underline {56.3} &
		\textbf{36.3} &
		45.8 &
		70.1 &
		\textbf{94.1} &
		\textbf{62.1} &
		\underline {73.5} &
		57.9 &
		\textbf{78.1} &
		47.2 &
		\textbf{40.3} &
		\textbf{79.4} &
		\underline {75.3} &
		\underline {63.0} &
		\textbf{54.1} &
		39.8 &
		\textbf{78.9} &
		\underline {97.8} &
		\textbf{59.7} \\
		SpectralX\tiny $*$SatMAE++ &
		ViT-L &
		11.8M &
		70.9 &
		70.6 &
		64.7 &
		36.6 &
		\textbf{18.7} &
		\textbf{20.6} &
		54.2 &
		33.5 &
		44.8 &
		\textbf{71.6} &
		\underline {93.9} &
		58.8 &
		\textbf{76.3} &
		\textbf{58.5} &
		75.4 &
		46.2 &
		39.1 &
		76.6 &
		\textbf{76.9} &
		\textbf{63.2} &
		49.8 &
		\textbf{55.3} &
		75.9 &
		\textbf{98.1} &
		\underline {59.6} \\ \hline
	\end{tabular}
	\end{adjustwidth}
{${\textbf{\dag}}$-fine-tuning full of the foundation model, $*$-fine-tuning part of the foundation model.}
\end{table*}
\begin{table}[]
	\setlength\tabcolsep{0.5pt}
	\caption{\label{table:DFC_miou}
	Performance comparison without domain gap on DFC2020. Bold are the best scores, underline are second ones. }
	\begin{adjustwidth}{-0.5cm}{-0.5cm}
	\begin{tabular}{cccccccccccc}
		\hline
		Method &
		Backbone &
		\begin{tabular}[c]{c}Trainable\\ params\end{tabular} &
		C1 &
		C2 &
		C3 &
		C4 &
		C5 &
		C6 &
		C7 &
		C8 &
		mIoU \\ \hline
		\multicolumn{12}{c}{\cellcolor[HTML]{E7E7E7}\textit{\textbf{Domain Adaptation/Generalization}}} \\ \hline
		SSC &
		ResNet101 &
		50.4M &
		79.3 &
		37.4 &
		51.4 &
		41.3 &
		62.4 &
		66.7 &
		25.1 &
		98.2 &
		57.7 \\
		DAEN &
		RefineNet-lw &
		48.7M &
		{\underline{83.7}} &
		44.3 &
		58.1 &
		47.5 &
		69.0 &
		{\underline{73.9}} &
		33.2 &
		\textbf{98.8} &
		63.5 \\
		DSTC &
		PVTv2-B1 &
		25.7M &
		80.6 &
		46.4 &
		57.0 &
		50.9 &
		68.8 &
		69.5 &
		38.6 &
		97.8 &
		63.4 \\ \hline
		\multicolumn{12}{c}{\cellcolor[HTML]{E7E7E7}\textit{\textbf{Remote Sensing/Spectral Foundation Models}}} \\ \hline
		ScaleMAE$^{\textbf{\dag}}$ &
		ViT-L &
		303.4M &
		79.3 &
		43.1 &
		54.0 &
		48.5 &
		66.1 &
		68.0 &
		37.8 &
		98.3 &
		61.9 \\
		SatMAE++$^{\textbf{\dag}}$ &
		ViT-L &
		305.5M &
		80.6 &
		44.0 &
		56.8 &
		50.8 &
		68.3 &
		68.9 &
		36.6 &
		98.2 &
		63.0 \\
		SpectralGPT+$^{\textbf{\dag}}$ &
		ViT-L &
		307.2M &
		82.9 &
		47.5 &
		57.6 &
		50.8 &
		69.2 &
		72.5 &
		39.2 &
		97.2 &
		64.9 \\
		HyperSIGMA$^{\textbf{\dag}}$ &
		ViT-L &
		382.5M &
		83.2 &
		47.9 &
		\underline{59.9} &
		52.6 &
		70.2 &
		72.0 &
		38.4 &
		98.5 &
		65.3 \\ \hline
		\multicolumn{12}{c}{\cellcolor[HTML]{E7E7E7}\textit{\textbf{Parameter-Efficient Fine-Tuning}}}\\ \hline
		LoRA\tiny $*$ScaleMAE &
		ViT-L &
		6.9M &
		82.9 &
		41.0 &
		59.8 &
		52.8 &
		70.3 &
		65.4 &
		{\underline{42.8}} &
		98.5 &
		64.2 \\
		LoRA\tiny $*$SatMAE++ &
		ViT-L &
		6.9M &
		81.6 &
		46.0 &
		57.7 &
		50.9 &
		68.6 &
		70.0 &
		41.3 &
		98.5 &
		64.3 \\
		SLR\tiny $*$ScaleMAE &
		ViT-L &
		7.3M &
		82.9 &
		\textbf{50.5} &
		53.9 &
		52.6 &
		{\underline{70.6}} &
		67.4 &
		42.2 &
		96.6 &
		64.6 \\
		SLR\tiny $*$SatMAE++ &
		ViT-L &
		7.3M &
		82.2 &
		48.6 &
		59.2 &
		52.8 &
		69.7 &
		71.1 &
		41.4 &
		98.6 &
		65.4 \\
		SpectralX\tiny $*$ScaleMAE &
		ViT-L &
		11.8M &
		\textbf{83.9} &
		{\underline{50.2}} &
		\textbf{61.4} &
		{\underline {53.4}} &
		\textbf{71.2} &
		72.7 &
		41.8 &
		{\underline {98.7}} &
		\textbf{66.7} \\
		SpectralX\tiny $*$SatMAE++ &
		ViT-L &
		11.8M &
		82.6 &
		47.1 &
		58.2 &
		\textbf{55.0} &
		68.5 &
		\textbf{74.7} &
		\textbf{44.6} &
		98.5 &
		{\underline {66.2}} \\ \hline
	\end{tabular}
	\end{adjustwidth}
{${\textbf{\dag}}$-fine-tuning full of the foundation model, $*$-fine-tuning part of the foundation model.}
\end{table}

\begin{table}[]
		\setlength\tabcolsep{0.5pt}
	\caption{\label{table:MTS12_miou}
		Performance comparison without domain gap on MTS12. Bold are the best scores, underline are second ones. }
	\begin{adjustwidth}{-0.5cm}{-0.5cm}
	\begin{tabular}{cccccccccccc}
		\hline
		Method &
		Backbone &
		\begin{tabular}[c]{c}Trainable\\ params\end{tabular}  &
		C1 &
		C2 &
		C3 &
		C4 &
		C5 &
		C6 &
		C7 &
		C8 &
		mIoU \\ \hline
		\multicolumn{12}{c}{\cellcolor[HTML]{E7E7E7}\textit{\textbf{Domain Adaptation/Generalization}}} \\ \hline
		SSC &
		ResNet101 &
		50.4M &
		44.8 &
		82.5 &
		42 &
		3.8 &
		0.0 &
		0.0 &
		27.6 &
		48.7 &
		31.2 \\
		DAEN &
	RefineNet-lw &
		48.7M &
		42.5 &
		85.7 &
		46.5 &
		8.3 &
		34.5 &
		0.0 &
		39.3 &
		43.9 &
		37.6 \\
		DSTC &
		PVTv2-B1 &
		25.7M &
		50.4 &
		83.0 &
		42.2 &
		14.1 &
		38.9 &
		\textbf{0.4} &
		37.8 &
		49.5 &
		39.5 \\ \hline
		\multicolumn{12}{c}{\cellcolor[HTML]{E7E7E7}\textit{\textbf{Remote Sensing/Spectral Foundation Models}}} \\ \hline
		ScaleMAE$^{\textbf{\dag}}$ &
		ViT-L &
		303.4M &
		46.1 &
		82.1 &
		39.8 &
		11.2 &
		33.1 &
		0.0 &
		31.8 &
		43.5 &
		35.9 \\
		SatMAE++$^{\textbf{\dag}}$ & ViT-L & 305.5M & 48.2 &
		81.6 &
		40.6 &
		9.8 &
		35.0 &
		0.0 &
		35.1 &
		\textbf{58.3} &
		38.6 \\
		SpectralGPT+$^{\textbf{\dag}}$ &
		ViT-L &
		307.2M &
		45.7 &
		72.5 &
		40.9 &
		12.5 &
		36.7 &
		{\underline {0.2}} &
		34.9 &
		44.2 &
		36.0 \\
		HyperSIGMA$^{\textbf{\dag}}$ &
		ViT-L &
		382.5M &
		53.5 & \textbf{87.1} & \textbf{48.8} & 14.3 & {\underline {47.3}} & 0.1 & \textbf{42.7} & 37.8 & 41.6 \\ \hline
		\multicolumn{12}{c}{\cellcolor[HTML]{E7E7E7}\textit{\textbf{Parameter-Efficient Fine-Tuning}}} \\ \hline
		LoRA\tiny $*$ScaleMAE &
		ViT-L &
		6.9M &
		\textbf{54.5} &
		79.8 &
		{\underline {47.7}} &
		13.4 &
		42.5 &
		0.0 &
		39.9 &
		52.2 &
		41.2 \\
		LoRA\tiny $*$SatMAE++ &
		ViT-L &
		6.9M &
		{\underline {54.0}} &
		81.9 &
		45.1 &
		\textbf{14.8} &
		42.6 &
		0.1 &
		{\underline {41.8}} &
		54.2 &
		41.8 \\
		SLR\tiny $*$ScaleMAE &
		ViT-L &
		7.3M &
		50.4 &
		82.2 &
		42.1 &
		{\underline {14.7}} &
		44.2 &
		0.1 &
		40.5 &
		53.4 &
		40.9 \\
		SLR\tiny $*$SatMAE++ &
		ViT-L &
		7.3M &
		51.4 &
		79.0 &
		44.9 &
		14.6 &
		\textbf{48.6} &
		0.3 &
		40.6 &
		{\underline {57.5}} &
		{\underline {42.1}} \\
		SpectralX\tiny $*$ScaleMAE &
		ViT-L &
		11.8M &
		52.9 &
		85.2 &
		45.8 &
		14.1 &
		41.5 &
		0.0 &
		39.0 &
		57.4 &
		42.0 \\
		SpectralX\tiny $*$SatMAE++ &
		ViT-L &
		11.8M &
		53.1 &
		{\underline {86.3}} &
		46.6 &
		14.0 &
		43.2 &
		0.0 &
		41.5 &
		55.1 &
		\textbf{42.5} \\\hline
	\end{tabular}
	\end{adjustwidth}
{${\textbf{\dag}}$-fine-tuning full of the foundation model, $*$-fine-tuning part of the foundation model.}
\end{table}

\begin{table}[]
	\setlength\tabcolsep{4.5pt}
	\caption{\label{table:WHUOHS_DG_miou} 
		Performance comparison with regional domain gap on WHUOHS. Bold are the best scores, underline are second ones. }
\begin{center}
	\begin{tabular}{ccccccc}
		\hline
		\multirow{2}{*}{Method}               & \multicolumn{3}{c}{WS→WD} & \multicolumn{3}{c}{WD→WS}   \\ \cline{2-7} 
		&
		mIoU &
		\multicolumn{1}{l}{m-F1} &
		\multicolumn{1}{l}{m-acc} &
		mIoU &
		\multicolumn{1}{l}{m-F1} &
		\multicolumn{1}{l}{m-acc} \\ \hline
		\multicolumn{7}{c}{\cellcolor[HTML]{E7E7E7}\textit{\textbf{Domain Adaptation/Generalization}}}                                      \\ \hline
		SSC                                   & 9.4     & 15.0   & 32.1   & 10.8 & 16.8 & 31.1          \\
		DAEN                                  & 11.9    & 17.9   & 35.6   & 11.5 & 18.2 & 37.7          \\
		DSTC                                  & 12.5    & 18.9   & 39.2   & 13.1 & 20.1 & 40.3          \\ \hline
		\multicolumn{7}{c}{\cellcolor[HTML]{E7E7E7}\textit{\textbf{Remote Sensing/Spectral Foundation Models}}}                                   \\ \hline
		ScaleMAE$^{\textbf{\dag}}$            & 12.0    & 18.3   & 38.8   & 11.3 & 17.9 & 37.1          \\
		SatMAE++$^{\textbf{\dag}}$            & 12.7    & 19.5   & 41.4   & 13.4 & 20.7 & 40.9  \\
		SpectralGPT+$^{\textbf{\dag}}$        & 11.9    & 18.2   & 35.8   & 12.4 & 18.9 & 39.1          \\
		HyperSIGMA$^{\textbf{\dag}}$           & 14.1    & 21.1   & 42.9   & 14.1 & 21.5 & \textbf{42.4}         \\ \hline
		\multicolumn{7}{c}{\cellcolor[HTML]{E7E7E7}\textit{\textbf{Parameter-Efficient Fine-Tuning}}}                                       \\ \hline
		LoRA\tiny $*$ScaleMAE & 13.1    & 20.3   & 42.8   & 12.9 & 19.9 & 40.1          \\
		LoRA\tiny $*$SatMAE++ & 12.7    & 20.0   & 39.9   & 11.3 & 17.5 & 37.4          \\
		SLR\tiny $*$ScaleMAE  & 13.5    & 20.8   & 43.5   & 13.3 & 20.4 & 40.6          \\
		SLR\tiny $*$SatMAE++  & 12.6    & 19.5   & 41.0   & 12.1 & 18.5 & 38.8          \\
		SpectralX\tiny $*$ScaleMAE &
		\textbf{15.6} &
		\textbf{23.2} &
		\textbf{45.5} &
		\textbf{14.9} &
		\textbf{22.7} &
		{\underline {41.8}} \\
		SpectralX\tiny $*$SatMAE++ &
		{\underline {14.2}} &
		{\underline {21.2}} &
		{\underline {44.5}} &
		{\underline {14.4}} &
		{\underline {22.0}} &
		41.1 \\\hline
	\end{tabular}
\end{center}
{${\textbf{\dag}}$-fine-tuning full of the foundation model, $*$-fine-tuning part of the foundation model.}
\end{table}

\begin{table*}[]
	\vspace{-2em}
	\caption{\label{table:DFC_DG_miou}
	Performance comparison with seasonal domain gap on DFC2020. Bold are the best scores, underline are second ones. }
		\begin{center}
	\begin{tabular}{ccccccccccccc}
		\hline
		\multirow{2}{*}{Method} &
		\multicolumn{3}{c}{Spr→S,A,W} &
		\multicolumn{3}{c}{Sum→S,A,W} &
		\multicolumn{3}{c}{Aut→S,S,W} &
		\multicolumn{3}{c}{Win→S,S,A} \\ \cline{2-13} 
		&
		mIoU &
		m-F1 &
		m-acc &
		mIoU &
		m-F1 &
		m-acc &
		mIoU &
		m-F1 &
		m-acc &
		mIoU &
		m-F1 &
		m-acc \\ \hline
		\multicolumn{13}{c}{\cellcolor[HTML]{E7E7E7}\textit{\textbf{Domain Adaptation/Generalization}}}  \\ \hline
		SSC &
		13.2 &
		21.3 &
		31.5 &
		12.5 &
		20.8 &
		29.8 &
		30.1 &
		40.0 &
		51.8 &
		24.1 &
		34.2 &
		52.0 \\
		DAEN &
		12.8 &
		21.3 &
		31.5 &
		11.8 &
		19.8 &
		26.4 &
		29.7 &
		38.5 &
		48.6 &
		25.1 &
		35.0 &
		53.4 \\
		DSTC &
		12.8 &
		21.1 &
		31.2 &
		13.3 &
		21.8 &
		26.6 &
		30.8 &
		40.6 &
		52.0 &
		26.4 &
		37.6 &
		54.7 \\ \hline
		\multicolumn{13}{c}{\cellcolor[HTML]{E7E7E7}\textit{\textbf{Remote Sensing/Spectral Foundation Models}}} \\ \hline
		ScaleMAE$^{\textbf{\dag}}$ &
		10.4 &
		17.6 &
		28.0 &
		10.8 &
		18.0 &
		25.3 &
		28.8 &
		38.2 &
		47.5 &
		24.4 &
		34.2 &
		52.9 \\
		SatMAE++$^{\textbf{\dag}}$ &
		11.2 &
		18.7 &
		27.9 &
		{\underline {14.2}} &
		{\underline {22.8}} &
		{\underline {31.1}} &
		29.8 &
		38.8 &
		51.0 &
		25.5 &
		35.1 &
		54.3 \\
		SpectralGPT+$^{\textbf{\dag}}$ &
		12.5 &
		19.8 &
		30.9 &
		13.7 &
		22.2 &
		30.7 &
		31.3 &
		41.7 &
		54.0 &
		26.1 &
		37.0 &
		54.4 \\
		HyperSIGMA$^{\textbf{\dag}}$ &
		11.2 &
		18.8 &
		28.4 &
		13.0 &
		21.3 &
		26.0 &
		30.6 &
		40.4 &
		54.7 &
		28.1 &
		38.2 &
		57.4  \\ \hline
		\multicolumn{13}{c}{\cellcolor[HTML]{E7E7E7}\textit{\textbf{Parameter-Efficient Fine-Tuning}}} \\ \hline
		LoRA\tiny $*$ScaleMAE &
		12.6 &
		20.8 &
		31.1 &
		12.4 &
		20.6 &
		29.9 &
		31.0 &
		41.4 &
		53.6 &
		27.2 &
		37.9 &
		56.9 \\
		LoRA\tiny $*$SatMAE++ &
		12.9 &
		21.1 &
		31.5 &
		13.5 &
		21.8 &
		30.7 &
		30.3 &
		40.4 &
		52.3 &
		27.5 &
		38.0 &
		56.5 \\
		SLR\tiny $*$ScaleMAE &
		{\underline {13.2}} &
		{\underline {21.5}} &
		31.8 &
		13.1 &
		21.0 &
		30.4 &
		30.2 &
		40.4 &
		52.2 &
		27.8 &
		38.3 &
		57.8 \\
		SLR\tiny $*$SatMAE++ &
		13.0 &
		21.4 &
		31.6 &
		14.0 &
		22.6 &
		30.8 &
		31.4 &
		41.9 &
		53.9 &
		{\underline {28.6}} &
		38.7 &
		58.1 \\
		SpectralX\tiny $*$ScaleMAE &
		\textbf{13.9} &
		\textbf{21.8} &
		\textbf{32.4} &
		\textbf{14.6} &
		\textbf{23.1} &
		\textbf{31.4} &
		{\underline {32.3}} &
		{\underline {42.2}} &
		{\underline {54.8}} &
		{\underline {28.6}} &
		{\underline {38.8}} &
		{\underline {58.5}} \\
		SpectralX\tiny $*$SatMAE++ &
		{\underline {13.2}} &
		{\underline {21.5}} &
		{\underline {31.9}} &
		13.9 &
		22.5 &
		30.9 &
		\textbf{32.8} &
		\textbf{43.0} &
		\textbf{55.0} &
		\textbf{30.6} &
		\textbf{40.2} &
		\textbf{61.1} \\ \hline
	\end{tabular}
	\end{center}
{${\textbf{\dag}}$-fine-tuning full of the foundation model, $*$-fine-tuning part of the foundation model.}
\end{table*}

\begin{table}[]
	\setlength\tabcolsep{4.5pt}
	\caption{\label{table:MTS12_DG_miou}
	Performance comparison with regional domain gap on MTS12. Bold are the best scores, underline are second ones. }
\begin{center}
	\begin{tabular}{ccccccc}
		\hline
		\multirow{2}{*}{Method}               & \multicolumn{3}{c}{N→S}        & \multicolumn{3}{c}{S→N} \\ \cline{2-7} 
		&
		mIoU &
		\multicolumn{1}{l}{m-F1} &
		\multicolumn{1}{l}{m-acc} &
		mIoU &
		\multicolumn{1}{l}{m-F1} &
		\multicolumn{1}{l}{m-acc} \\ \hline
		\multicolumn{7}{c}{\cellcolor[HTML]{E7E7E7}\textit{\textbf{Domain Adaptation/Generalization}}}                                             \\ \hline
		SSC                                   & 25.7       & 34.8       & 76.6 & 21.8   & 29.3   & 71.5  \\
		DAEN                                  & 23.8       & 32.0       & 76.7 & 23.0   & 32.0   & 69.2  \\
		DSTC                                  & 28.7       & 38.5       & 80.6 & 26.5   & 35.7   & 73.0  \\ \hline
		\multicolumn{7}{c}{\cellcolor[HTML]{E7E7E7}\textit{\textbf{Remote Sensing/Spectral Foundation Models}}}                                    \\ \hline
		ScaleMAE$^{\textbf{\dag}}$            & 25.7       & 34.6       & 78.6 & 23.0   & 31.9   & 69.5  \\
		SatMAE++$^{\textbf{\dag}}$            & 25.7       & 34.4       & 78.9 & 23.8   & 32.9   & 69.9  \\
		SpectralGPT+$^{\textbf{\dag}}$        & 25.0       & 33.9       & 77.9 & 24.8   & 33.7   & 72.4  \\
		HyperSIGMA$^{\textbf{\dag}}$          & {\underline {29.7}} & {\underline {39.3}} & 81.8 & 26.4   & 35.4   & 73.3  \\ \hline
		\multicolumn{7}{c}{\cellcolor[HTML]{E7E7E7}\textit{\textbf{Parameter-Efficient Fine-Tuning}}}                                              \\ \hline
		LoRA\tiny $*$ScaleMAE & 27.6       & 37.9       & 79.8 & 25.8   & 34.9   & 73.1  \\
		LoRA\tiny $*$SatMAE++ & 28.2       & 38.2       & 80.4 & 26.3   & 35.7   & 73.8  \\
		SLR\tiny $*$ScaleMAE  & 26.7       & 37.1       & 79.1 & 26.1   & 35.3   & 73.5  \\
		SLR\tiny $*$SatMAE++  & 27.1       & 37.8       & 79.5 & 26.6   & 36.0   & 73.9  \\
		SpectralX\tiny $*$ScaleMAE &
		{\underline {29.7}} &
		{\underline {39.3}} &
		81.4 &
		\textbf{27.5} &
		\textbf{37.0} &
		\textbf{74.6} \\
		SpectralX\tiny $*$SatMAE++ &
		\textbf{30.5} &
		\textbf{40.2} &
		\textbf{82.6} &
		{\underline {27.1}} &
		{\underline {36.6}} &
		{\underline {74.3}} \\ \hline
	\end{tabular}
\end{center}
{${\textbf{\dag}}$-fine-tuning full of the foundation model, $*$-fine-tuning part of the foundation model.}
\end{table}

We categorize all tasks in the three datasets based on the presence or absence of domain gaps. The task abbreviations, the number of training and test images, image sizes, and the number of classes are shown in Table \ref{table:data}. To explore the performance changes of existing methods under various domain gaps, with or without domain gaps, and to demonstrate the generalization capability of SpectralX, we conducted extensive experiments by comparing it with a series of representative models on the spectral domain generalization benchmark constructed from these three datasets. The comparison methods in the experiment include domain adaptation/generalization methods, Spatial Structure Constraints (SSC) \cite{10418852}, Domain Adaptive Energy-based Network (DAEN) \cite{10735117}, Dual-stage Spectral Supertoken Classifier (DSTC) \cite{liu2024dual}, Remote Sensing/Spectral Foundation Models, Scale-MAE \cite{reed2023scale}, SatMAE++ \cite{noman2024rethinking}, SpectralGPT \cite{hong2024spectralgpt}, HyperSIGMA \cite{wang2024hypersigma}.  and Parameter-Efficient Fine-Tuning methods, Low-Rank Adaptation (LoRA) \cite{hu2022lora}, Scaled Low-Rank (SLR) \cite{scheibenreif2024parameter}.

\subsection{Experimental Data}

\textbf{WHUOHS dataset (HSI)}:
The WHUOHS dataset consists of more than 40 different locations in China, acquired by the Orbita hyperspectral micro-nano satellite. This dataset features extensive geographical distribution, large spatial coverage, and a widely used classification system. For details on the wavelength range, one can refer to \cite{li2022whu}. The cross-domain task settings of WHUOHS dataset are shown in Table \ref{table:data}.

\textbf{DFC2020 dataset (MSI)}: The DFC2020 dataset, acquired by the Sentinel-2 satellite, consists of seven different locations worldwide, including Khabarovsk, Russia; Mumbai, India; Kippa Ring, Australia; Mexico City, Mexico; Bandar Anzali, Iran; Black Forest, Germany; and Cape Town, South Africa. The cross-domain task settings of DFC2020 dataset are shown in Table \ref{table:data}.

\textbf{MTS12 dataset (Multi-Temporal MSI)}: The MTS12 dataset consists of Sentinel-2 time series data from January to December 2019, covering the entire territory of Slovenia. The dataset divides Slovenia into southern and northern regions, encompassing 8 land cover classes. The cross-domain task settings of MTS12 dataset are shown in Table \ref{table:data}.

\begin{figure*}[tp] \small
	\begin{center}
		\centering
		\epsfig{width=2.3\figurewidth,file=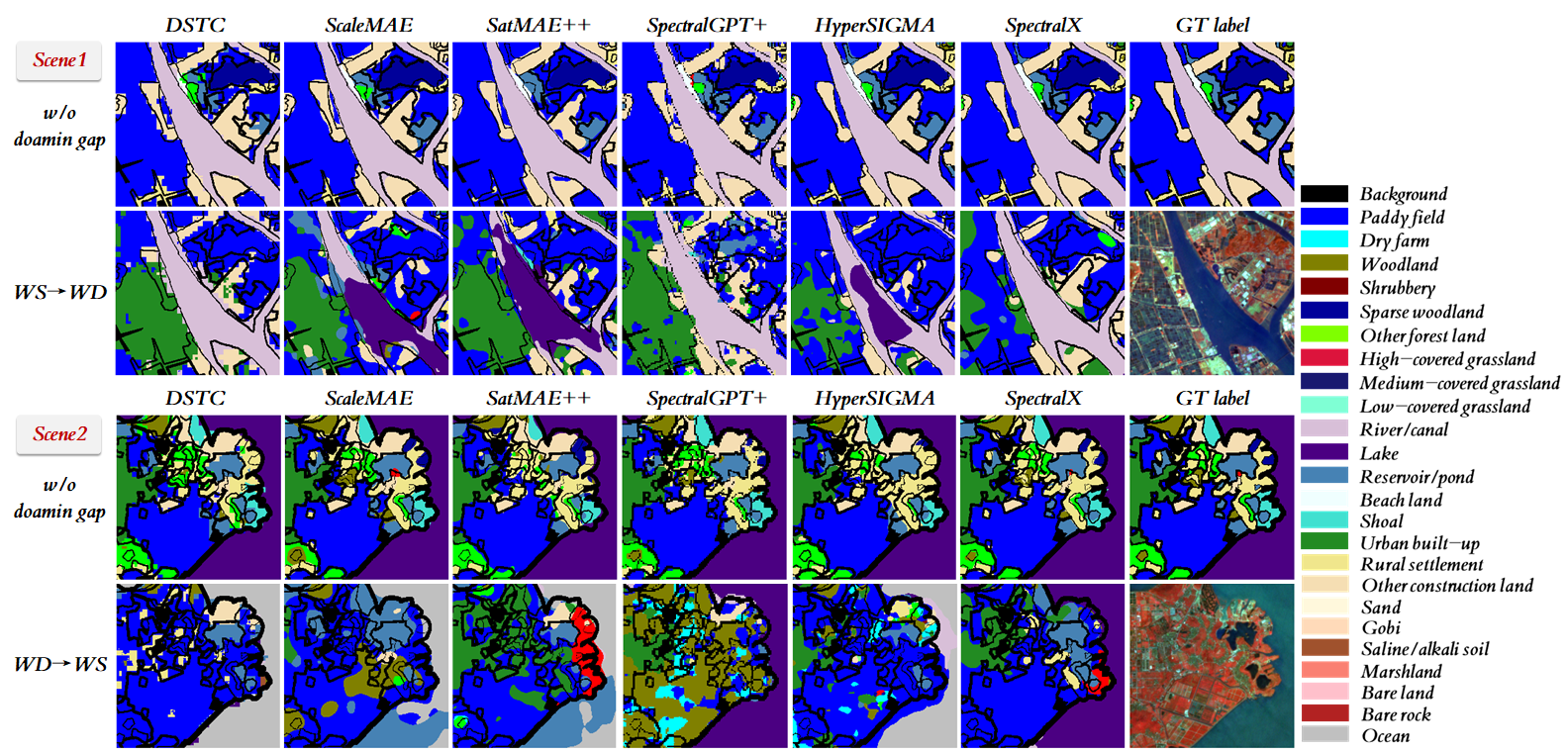}
		\caption{\label{fig:WHUOHS_segmap}
			Visualization of predictive segmentation maps without and with domain gap in WHUOHS.}
	\end{center}
	\vspace{-2em}
\end{figure*}

\begin{figure*}[tp] \small
	\begin{center}
		\centering
		\epsfig{width=2.3\figurewidth,file=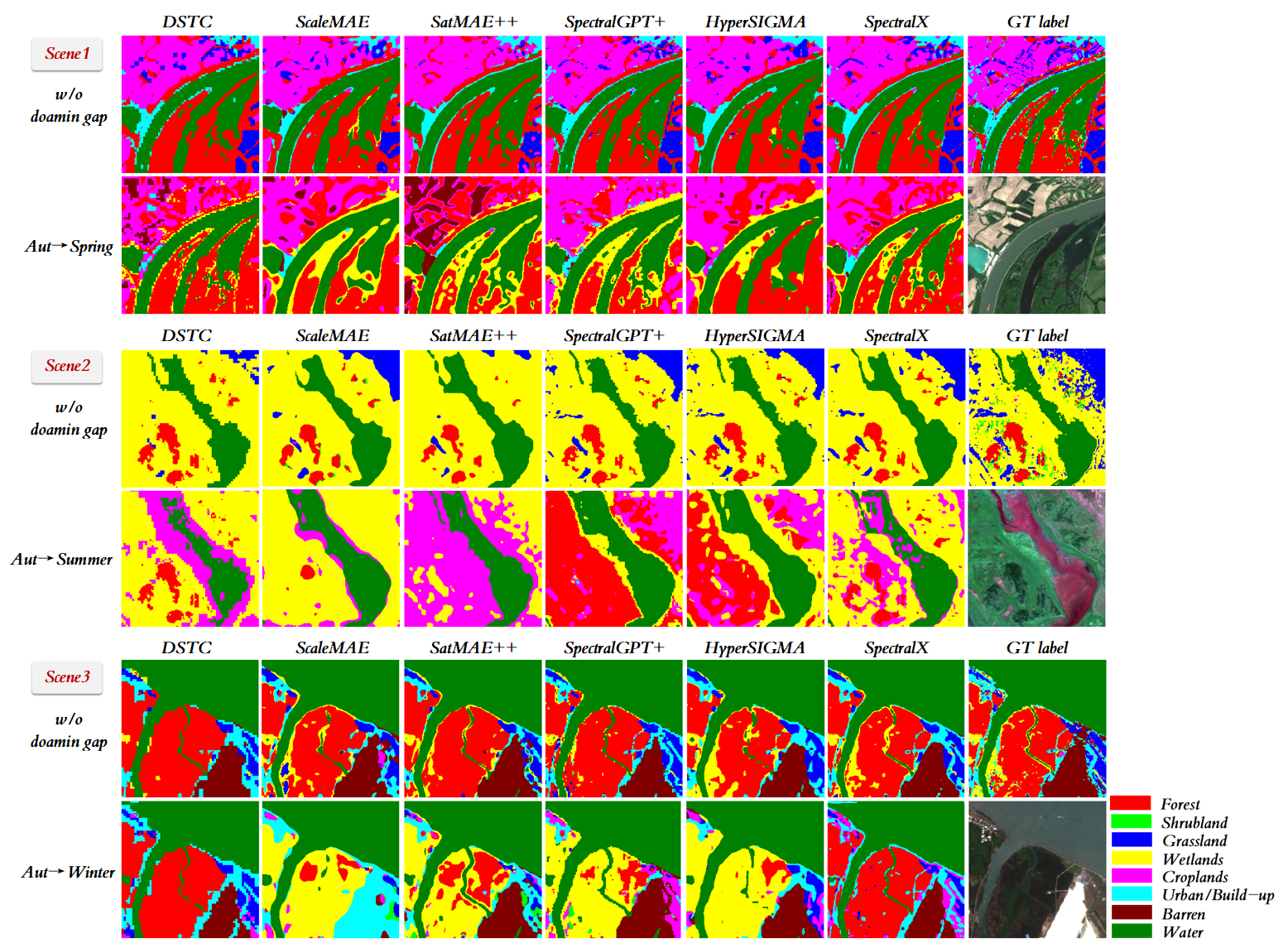}
		\caption{\label{fig:DFC_segmap}
			Visualization of predictive segmentation maps without and with domain gap in DFC2020.}
	\end{center}
	\vspace{-2em}
\end{figure*}

\subsection{Ablation study}

We conducted ablation studies on the WHUOHS and DFC2020 datasets, analyzing the performance of each module under both w/o domain gap and w/ domain gap conditions, as shown in Table \ref{table:Ablation}. Using Scale-MAE as the backbone, we describe the various ablation settings in our experiments as follows:

\begin{itemize}
	
	\item Full fine-tuning: The patch embedding layer is reset according to the spectral band count of the input data, and all parameters of Scale-MAE are fine-tuned directly.
	
	\item Freeze: All parameters of Scale-MAE are frozen, with only the patch embedding layer being fine-tuned.
		
	\item w/o stage1: Without spectral modality adaptation training, we analyze the contributions of the HyperT, AoMoA, and Are-adapter modules, where + w/ indicates adding the current module on top of the previous one.
	
	\item w/ stage1: With spectral modality adaptation training enabled, we evaluate the performance gains of the aforementioned modules.
	
\end{itemize}

SpectralX extends RSFMs to spectral modalities solely through task-oriented generalization training without stage1. Compared to freeze Scale-MAE, the HyperT demonstrates superior capability in extracting spatial-spectral information over conventional patch embedding. By incrementally incorporating AoMoA and Are-adapter, we observe that there are different degrees of gain in the w/o or w/ domain gap. Notably, when AoMoA is applied to multiple transformer blocks for fine-tuning, it achieves approximately 2\% improvement over full fine-tuning on both WHUOHS and DFC2020 datasets in w/o domain generalization conditions. The incorporation of stage1 provides significant performance enhancements across all three key components, delivering consistent improvements of 2\%-3\% compared to configurations without stage1. This demonstrates that spectral modality adaptation serves as an effective initialization process, properly conditioning the RSFMs for spectral modality characteristics and thereby enabling more effective fine-tuning for subsequent downstream tasks.

\subsection{Performance on segmentation without domain gap}

In the introduction, we posed the question: \textit{\textbf{How can RSFMs be adapted to spectral modalities and achieve interpretation performance comparable to that of SpectralFMs, without requiring extensive spectral pretraining?}} In this section of the experiment, without considering domain gaps, we first validate the modality transfer performance of SpectralX. Tables \ref{table:WHUOHS_miou}-\ref{table:MTS12_miou} report the class-wise IoU and mIoU on three datasets. The results and corresponding analysis are discussed as follows.

\begin{itemize}
	
	\item Existing domain adaptation/generalization methods show little difference in segmentation task performance compared to most RS/Spectral FMs. RS/Spectral FMs do not achieve significant performance improvements in segmentation tasks due to pre-training based on optical or spectral modalities, with HyperSIGMA performing the best across the three datasets. This is primarily attributed to HyperSIGMA's pre-training on 450,000 HSI images from around the world, which provides it with rich spectral modality knowledge.
	
	\item The two PEFT methods, LoRA and SLR, fine-tune a small number of parameters in the MLP and MSA layers of RSFMs to adapt them to spectral modalities. Compared to fully fine-tuning Scale-MAE and SatMAE++ in Tables \ref{table:DFC_miou}-\ref{table:MTS12_miou}, LoRA and SLR bring more performance improvements to Scale-MAE and SatMAE++ on the MSI datasets (DFC2020 and MTS12) in a more economical manner. Particularly, on the MTS12 dataset, SLR{\tiny $*$SatMAE++} improves by 3.5\% compared to SatMAE++, and LoRA{\tiny $*$ScaleMAE} improves by 5.3\% compared to ScaleMAE. However, LoRA and SLR cannot adapt to hyperspectral modalities, and their performance on the WHUOHS dataset is far inferior to fully fine-tuning Scale-MAE and SatMAE++.
	
	\item SpectralX demonstrates the best performance compared to other methods, particularly on the WHUOHS dataset, where SpectralX{\tiny $*$ScaleMAE} achieves a 2\% improvement in mIoU over HyperSIGMA's 57.7\%. Compared to LoRA and SLR, SpectralX not only enables RSFMs to effectively transfer from optical to multispectral modalities but, more importantly, excels in adapting to hyperspectral modalities. When fine-tuning ScaleMAE, SpectralX{\tiny $*$ScaleMAE} achieves a 7.4\% mIoU improvement over SLR{\tiny $*$ScaleMAE}'s 52.4\%. The outstanding spectral modality adaptation capability is primarily due to the decoupling of spatial and spectral attributes, as well as the design of the HyperT and AoMoA to progressively achieve attribute perception and attribute matching.
\end{itemize}



\subsection{Performance on segmentation with domain gap}

Building on SpectralX's strong spectral modality transfer capability, we further validate its interpretation performance for spectral images with different domain gaps. Tables \ref{table:WHUOHS_DG_miou}-\ref{table:MTS12_DG_miou} report the mIoU, m-F1, and m-acc of all methods under different domain generalization task settings on the three datasets. The results and corresponding analysis are discussed as follows.

\begin{figure*}[tp] \small
		\vspace{-1em}
	\begin{center}
		\centering
		\epsfig{width=2\figurewidth,file=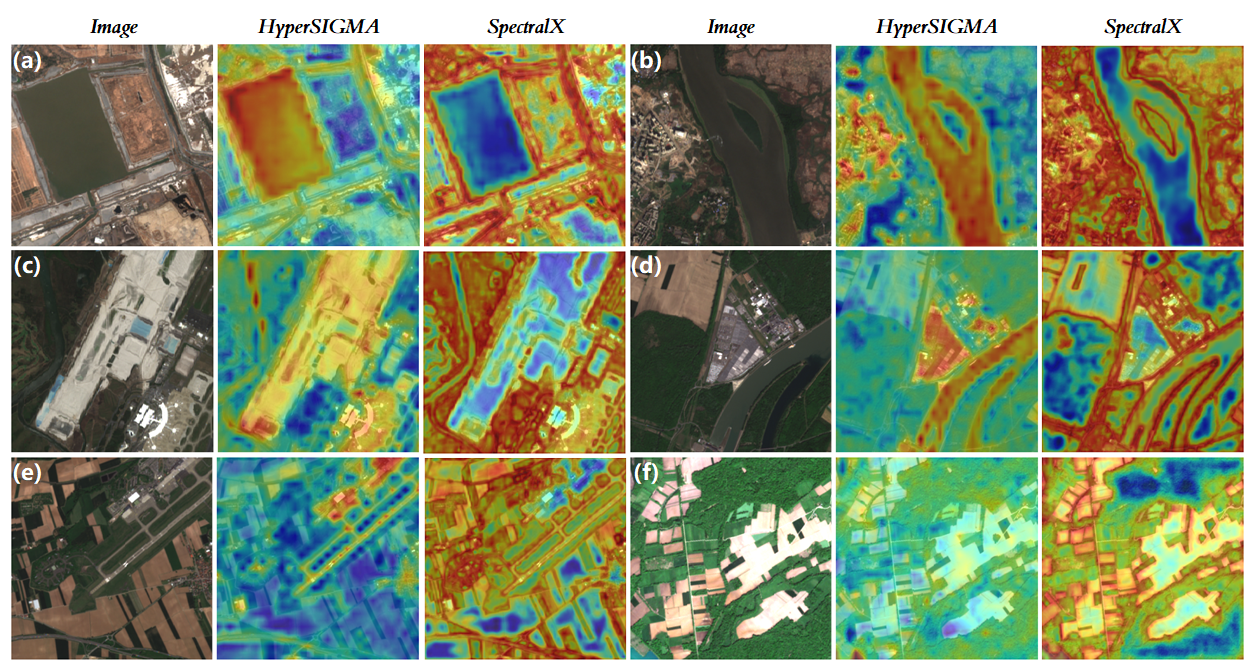}
		\caption{\label{fig:fea_DFC}
			Visualization of feature maps extracted by HyperSIGMA and SpectralX on DFC2020.}
	\end{center}
	\vspace{-2em}
\end{figure*}

\begin{itemize}
	
	\item DSTC outperforms other domain adaptation/generalization methods in most domain generalization tasks across the three datasets, demonstrating superior domain generalization interpretation performance compared to RSFMs. As shown in Table \ref{table:MTS12_DG_miou}, it significantly outperforms Scale-MAE and SatMAE++ on the MTS12 dataset, with an mIoU improvement of approximately 3\%. Combined with the experimental results without domain gaps, it can be concluded that although RSFMs can adapt to spectral modalities through full fine-tuning, their performance in downstream tasks with domain gaps is notably inferior to that of small-scale models specifically designed for domain generalization tasks.
	
	\item HyperSIGMA achieves the best performance in most domain generalization tasks compared to DSTC and other RS/Spectral FMs methods, particularly outperforming DSTC by 1.6\% mIoU and 1\% mIoU in the WS→WD and WD→WS tasks on WHUOHS. This indicates that incorporating large-scale unlabeled HSI into pre-training indeed enables superior generalization capabilities when facing regional/seasonal domain gaps in spectral images.
	
	\item In the six domain generalization tasks on DFC2020 and MTS12 (excluding Sum→S,A,W), LoRA and SLR effectively enhance the generalization performance of Scale-MAE and SatMAE++ on spectral modalities. The most significant improvements are observed in the Win→S,S,A task, where SLR{\tiny $*$ScaleMAE} improves by 3.4\% compared to Scale-MAE, and SLR{\tiny $*$SatMAE++} improves by 3.1\% compared to SatMAE++. In the S→N task, SLR{\tiny $*$ScaleMAE} improves by 3.1\% compared to Scale-MAE, and SLR{\tiny $*$SatMAE++} improves by 2.8\% compared to SatMAE++. However, as shown in Table \ref{table:WHUOHS_DG_miou}, LoRA and SLR not only fail to improve the performance of RSFMs but also degrade the performance of SatMAE++ in the WD→WS task. LoRA{\tiny $*$SatMAE++} decreases by -2.1\%, and SLR{\tiny $*$SatMAE++} decreases by -1.3\%.
	
	\item In the eight domain generalization tasks, SpectralX achieves the best generalization performance. In the WS→WD task, SpectralX{\tiny $*$ScaleMAE} improves by 1.5\% compared to HyperSIGMA, and in the Win→S,S,A task, SpectralX{\tiny $*$SatMAE++} improves by 2\% compared to SLR {\tiny $*$SatMAE++}. Additionally, compared to fully fine-tuning Scale-MAE and SatMAE++, SpectralX provides a more economical and effective fine-tuning approach, resulting in varying degrees of improvement in domain generalization interpretation performance. Combined with the Are-adapter, our fine-tuning process achieves task-oriented customized adjustments, focusing more on spatial and spectral attributes similar to the source domain when handling unseen regions/seasons, which helps mitigate domain gaps.
\end{itemize}

\subsection{Visualization of segmentation maps w/o or w/ domain gap}

Figs.~\ref{fig:WHUOHS_segmap}-\ref{fig:DFC_segmap} present the segmentation maps predicted by multiple methods under scenarios with and without domain gaps, visually demonstrating the impact of domain gaps on the models. These include one domain generalization tasks each from WHUOHS and one domain generalization task from DFC2020, specifically Aut→S,S,W (the visualization of segmentation maps and the analysis for MTS12 are provided in the supplemental materials). The visualization results are discussed as follows.

\begin{itemize}
	
	\item In the absence of domain gaps, most methods can accurately identify land cover classes within the scene, with the predicted segmentation maps closely matching the ground truth. For example, in Scene1 of Fig.~\ref{fig:WHUOHS_segmap}, land cover classes such as Paddy field, Medium-covered grassland, and Other construction land, as well as Woodland and Other forest land in Scene2, which exhibit similar spatial and spectral characteristics, are correctly identified. Similarly, in Fig.~\ref{fig:DFC_segmap}, Forest and Wetlands.
	
	\item However, all methods exhibit varying degrees of misidentification due to regional or seasonal domain gaps. For example, in the WHUOHS dataset (Fig.~\ref{fig:WHUOHS_segmap}) with regional domain gaps (WS→WD and WD→WS), the following errors are observed: In Scene1, ScaleMAE, SatMAE++, and HyperSIGMA misclassifies parts of River/canal as Lake, while all methods confuse Paddy field with Urban built-up. In Scene2, except for SpectralGPT+ and SpectralX, all other methods misclassify Lake as Ocean, whereas SpectralGPT+ incorrectly predicts large patches of Paddy field as Woodland. This indicates that when interpreting unseen regions, most methods tend to rely on the land cover distribution observed during training, leading to misidentification even for classes with obvious differences in land cover attributes. This further underscores the challenge of DG, where models struggle to account for spatial heterogeneity in land cover patterns.
	
	\item As shown in Fig.~\ref{fig:DFC_segmap}, when a model trained on autumn spectral data is directly applied to any season in the spring, summer and winter, Forest, Wetlands, and Croplands are frequently misidentification. SatMAE++, SpectralGPT+, and HyperSIGMA incorrectly identified large areas of Wetlands in Scene2 as Forest and Croplands, while misclassifying Forest as Wetlands in Scene3. SpectralX performed relatively better across all three seasons. Forest, Wetlands, and Croplands are all season-sensitive vegetation land cover types, and their growth states in autumn significantly differ from those in other seasons. Therefore, seasonal domain gaps introduce substantial misleading biases, and existing RS/SpectralFMs struggle to adapt to seasonal variations. This highlights the challenge of domain generalization in remote sensing, where models trained on a single season fail to generalize effectively to unseen seasonal conditions.

\end{itemize}

\subsection{Visualization of feature maps}

One of the main purposes of SpectralX is to be oriented towards downstream tasks, continuously perceiving relevant regions from spatial and spectral semantic features layer by layer, thereby achieving customized adjustments for downstream tasks. Fig.~\ref{fig:fea_DFC} shows the feature maps obtained by HyperSIGMA and SpectralX on the DFC2020 dataset (the visualization of feature maps of WHUOHS and MTS12 are provided in the supplemental materials). These features are derived from the final layer output of the backbone and represent feature maps rather than heatmaps. Therefore, we focus on the semantic boundaries in the feature maps rather than the color intensity.

The visualizations encompass a variety of scenes, including ports, farmlands, rivers, suburbs, and more. It is evident that SpectralX, compared to HyperSIGMA, can learn more discriminative semantic information about ground objects. For example, in the suburban scene Fig.~\ref{fig:fea_DFC}(f), HyperSIGMA can only focus on individual building areas, while SpectralX clearly delineates the semantic distinctions between building areas, roads, and farmlands. Similarly, Figs.~\ref{fig:fea_DFC}(b) and (d), SpectralX accurately depicts the semantic boundaries of different ground objects. Furthermore, observing Figs.~\ref{fig:fea_DFC}(a), (c), (d), and (e), it can be seen that SpectralX is capable of capturing high-quality semantics. Although DFC2020 consists of 10m low-resolution satellite images, complex road networks and airport layouts are well captured by SpectralX.

\section{Conclusions}
\label{sec:conclusions}
In this paper, the SpectralX is proposed, a parameter-efficient fine-tuning framework that extends Remote Sensing Foundation Models (RSFMs) built for optical modalities to spectral modalities while enhancing their domain generalization capabilities. SpectralX consists of three key stages: spectral modality adaptation (stage1), task-oriented generalization training (stage2), and unseen scenes interpretation (stage3). In stage1, we design a Hyper tokenizer (HyperT) tailored for spectral image to extract attribute tokens, along with an Attribute-oriented Mixture of Adapter (AoMoA) featuring a flexible routing scheme to dynamically fine-tune RSFMs. This stage mitigates the modality gap through reconstruction training, we further reduce the task gap by introducing an Attribute-refined Adapter (Are-adapter) in stage2, which progressively refines spatial and spectral attribute knowledge beneficial for downstream tasks, enabling customized adjustment for RSFMs. After training in these two stages, SpectralX is deployed for interpreting unseen scenes. SpectralX achieves cost-effective fine-tuning without relying on large-scale spectral pretraining, yet delivers strong generalization on spectral image. Extensive experiments on eight domain generalization tasks across three datasets demonstrate that SpectralX effectively adapts to various spectral images and outperforms existing methods in cross-domain tasks.

\bibliographystyle{IEEEtran}
\bibliography{bibfile_zyx}

\end{document}